\documentclass{article}

\usepackage[preprint, nonatbib]{neurips_2024}

\usepackage[utf8]{inputenc} 
\usepackage[T1]{fontenc}    
\usepackage{hyperref}       
\usepackage{url}            
\usepackage{booktabs}       
\usepackage{amsfonts}       
\usepackage{nicefrac}       
\usepackage{microtype}      
\usepackage{xcolor}         

\usepackage{algorithm}
\usepackage{algorithmic}

\usepackage{amsmath}
\usepackage{amssymb}
\usepackage{mathtools}
\usepackage{amsthm}

\usepackage{multirow}
\usepackage{wrapfig}

\usepackage{ulem}

\theoremstyle{plain}
\newtheorem{theorem}{Theorem}[section]

\theoremstyle{definition}
\newtheorem{definition}[theorem]{Definition}

\theoremstyle{remark}

\title{SE3Set: Harnessing equivariant hypergraph neural networks for molecular representation learning}

\author{%
  Hongfei Wu\thanks{These authors did this work during an internship at Microsoft Research AI4Science.} \\
  College of Chemistry and Molecular Engineering, \\
  Peking University, Beijing, 100871, China\\
  \And
  Lijun Wu \\
  Microsoft Research AI4Science, \\
  Beijing, 100084, China \\
  \AND
  Guoqing Liu \\
  Microsoft Research AI4Science, \\
  Beijing, 100084, China \\
  \And
  Zhirong Liu \\
  College of Chemistry and Molecular Engineering, \\
  Peking University, Beijing, 100871, China\\
  \texttt{LiuZhiRong@pku.edu.cn} \\
  \And
  Bin Shao \\
  Microsoft Research AI4Science, \\
  Beijing, 100084, China \\
  \texttt{binshao@microsoft.com} \\
  \And
  Zun Wang \\
  Microsoft Research AI4Science, \\
  Beijing, 100084, China \\
  \texttt{zunwang@microsoft.com}
}

\begin{document}

\maketitle

\begin{abstract}
In this paper, we develop SE3Set, an SE(3) equivariant hypergraph neural network architecture tailored for advanced molecular representation learning. Hypergraphs are not merely an extension of traditional graphs; they are pivotal for modeling high-order relationships, a capability that conventional equivariant graph-based methods lack due to their inherent limitations in representing intricate many-body interactions. To achieve this, we first construct hypergraphs via proposing a new fragmentation method that considers both chemical and three-dimensional spatial information of molecular system. We then design SE3Set, which incorporates equivariance into the hypergragh neural network. This ensures that the learned molecular representations are invariant to spatial transformations, thereby providing robustness essential for accurate prediction of molecular properties. SE3Set has shown performance on par with state-of-the-art (SOTA) models for small molecule datasets like QM9 and MD17. It excels on the MD22 dataset, achieving a notable improvement of approximately 20\% in accuracy across all molecules, which highlights the prevalence of complex many-body interactions in larger molecules. This exceptional performance of SE3Set across diverse molecular structures underscores its transformative potential in computational chemistry, offering a route to more accurate and physically nuanced modeling.
\end{abstract}

\section{Introduction}
Molecular representation~\cite{mathews2012molecular, david2020molecular, wigh2022review} is pivotal for cheminformatics~\cite{fourches2010trust}, impacting the prediction of molecular properties in drug discovery and material science. Traditional descriptors like fingerprints capture basic structural and energetic aspects of molecules by considering mainly one- and two-body interactions. However, they often miss complex electronic correlations and collective behaviors important for understanding phenomena such as chemical reactivity and protein folding. To address this, advanced methods that include many-body interactions are crucial for a more comprehensive molecular characterization. These methods enhance the predictive capabilities of computational models by more accurately reflecting the intricate dynamics and properties of molecules, which are essential for a deeper understanding of their functionality and reactivity in cheminformatics.

Graph neural networks (GNNs)~\cite{zhou2020graph, wu2020comprehensive} a foundational tool for representing structured data in molecular sciences molecular sciences with atoms as nodes and chemical bonds as edges, respectively. GNN models excel in tasks ranging from property prediction to reaction simulation~\cite{do2019graph, xiong2021graph, reiser2022graph}. GNNs can capture higher-order molecular interactions through message passing~\cite{gilmer2017neural} but face overfitting and inefficiency challenges~\cite{godwin2021simple, rusch2023survey}. Architectural improvements in GNNs facilitate the modeling of complex interactions, overcoming some limitations of deep networks~\cite{gasteiger2019directional, schutt2021equivariant, batzner2022}. Advances demonstrate the potential of architectural enhancements in GNNs to represent complex interactions~\cite{gasteiger2020fast, gasteiger2021gemnet, tholke2021equivariant, batatia2022mace, musaelian2023learning, wang2024enhancing}, but efficiently integrating many-body interactions into these networks is an ongoing challenge~\cite{wang2023efficiently}.

To address the complexities of many-body interactions in molecular systems, hypergraphs offer a compelling alternative to complex GNN architectures. Hypergraphs, with hyperedges connecting multiple vertices, can naturally represent many-body phenomena like electronic delocalization and collective vibrations. This allows for a more accurate modeling of molecular intricacies beyond the limitations of traditional graphs. Integrating hypergraphs with machine learning, particularly through Hypergraph Neural Networks (HGNNs), is an emerging research area. HGNNs manage the flow of information across hyperedges, capturing complex multi-atom interactions and enriching molecular representations. This technique promises to balance model expressiveness with computational efficiency. By innately encoding many-body interactions, HGNNs stand to significantly advance cheminformatics, offering a new approach to molecular property prediction and simulation that resonates with the actual behavior of chemical systems.

In this work, we introduce SE3Set, an innovative approach that enhances traditional GNNs by exploiting hypergraphs for modeling many-body interactions, while ensuring SE(3) equivariant representations that remain consistent regardless of molecular orientation. Our key contributions are:
\begin{itemize}
\item A new fragmentation method for hypergraph construction that seamlessly integrates 2D chemical and 3D spatial information, enriching the molecular structure representation.
\item The deployment of hypergraph neural networks to capture many-body interactions, providing a deeper insight into molecular behavior that surpasses conventional pairwise modeling.
\item The incorporation of SE(3) equivariance within our hypergraph framework, guaranteeing orientation-independent molecular representations.
\item SE3Set underwent a comprehensive benchmarking process, exhibiting comparable outcomes to state-of-the-art (SOTA) models on small molecule datasets QM9 and MD17. It demonstrated exceptional performance on the larger molecule dataset MD22, where higher-order interactions are more evident, surpassing SOTA models with a significant reduction in mean absolute errors (MAEs) by an average of roughly 20\%. This confirms SE3Set's efficacy in capturing the complexity of molecular representations.
\end{itemize}
These advances establish SE3Set as a formidable tool for molecular representation learning, with implications for computational chemistry and beyond.

\section{Related works}
\subsection{Graph neural networks}
Message passing neural networks (MPNNs), a class of graph neural networks, are essential for learning node features by transmitting information along graph edges, a process crucial for interpreting structured data like molecules~\cite{gilmer2017neural}. Equivariant GNNs are especially important for molecular modeling. They adopt either group representation methods, aligning architectures to symmetry groups for improved interaction modeling~\cite{thomas2018tensor, anderson2019cormorant, fuchs2020se, batzner2022, liao2022equiformer, liao2023equiformerv2, musaelian2023learning}, or direction-based methods that incorporate spatial information for accurate molecular representations~\cite{schutt2017schnet, schutt2018schnet, coors2018spherenet, gasteiger2019directional, gasteiger2020fast, schutt2021equivariant, tholke2021equivariant, gasteiger2021gemnet, wang2022comenet, wang2024enhancing} and have been engineered to handle intricate up to five-body interactions~\cite{wang2023efficiently}.

\subsection{Hypergraph neural networks}
Hypergraph Neural Networks (HGNNs) enhance GNNs by incorporating multi-node hyperedges, better capturing complexity in data from various domains. They advance GNNs' binary interactions with methods like clique expansion for compatibility with existing algorithms~\cite{agarwal2005beyond, zhou2006learning} and employ tensor techniques for improved hypergraph-based feature learning~\cite{li2013z, pearson2014spectral, benson2017spacey, chien2021landing, tudisco2021nonlinear}. While equivariant HGNNs adeptly handle node permutations, preserving data symmetries~\cite{kim2021transformers, kim2022equivariant}, they often miss 3D spatial transformations, crucial for physical system modeling. In computational chemistry, hypergraph algorithms simulate complex behaviors and optimize molecules through hypergraph grammar~\cite{cui2023hyper, tavakoli2022rxn, kajino2019molecular}, providing multidimensional insights into molecular structures~\cite{nachmani2020molecule, chen2021hypergraph, chen2023molecular}. Despite their promise, these methods still face hurdles in integrating spatial information effectively.

\subsection{Fragmentation methods}
Fragmentation methods break down complex molecules for simpler {\it ab initio} QM computations of properties, later combining these for a holistic view~\cite{gordon2012fragmentation, collins2015energy}. Leveraging the localized nature of chemical reactions, these techniques aim for scalable algorithms suitable for large molecule analysis. While instrumental in computational pretraining~\cite{du2021hypergraph, kim2022contrastive, luong2023fragment}, they typically neglect the fusion of 2D structural with 3D spatial data. Hence, we advocate for a refined fragmentation approach that merges chemical properties with spatial context, potentially advancing hypergraph-based chemical modeling.

\section{Preliminaries}
\subsection{Equivariance}
Consider a function $\mathcal{L}$ that maps inputs from space $\mathcal{X}$ to outputs in space $\mathcal{Y}$. $\mathcal{L}$ is called $G$-equivariant if it preserves the symmetry of a group $G$ across mappings, meaning for each $g \in G$, we have:
\begin{equation}
\mathcal{L} \circ D^{\mathcal{X}}(g) = D^{\mathcal{Y}}(g) \circ \mathcal{L},
\end{equation}
where $D^{\mathcal{X}}$ represents the group $G$'s action on $\mathcal{X}$. This ensures that the function $\mathcal{L}$ reflects changes made to inputs by $G$ in its outputs.

\subsection{Hypergraph}
Hypergraphs elegantly capture the essence of higher-order interactions among multiple entities, making them an invaluable tool for representing complex relational data. Let $\mathcal{G} = (V, E)$ be a hypergraph with $N$ vertices and $M$ hyperedges, where $V$ represents a set of nodes and $E$ is a set of hyperedges. Distinguishing itself from a traditional graph, a hyperedge can encompass multiple nodes, not limited to two, i.e. each hyperedge $e \in E$ is a non-empty subset of $V$.

\subsection{AllSet}
The AllSet framework~\cite{chien2021you}, an advanced HGNN model, addresses heuristic propagation rule limitations in HGNNs by integrating Deep Sets~\cite{zaheer2017deep} and Set Transformers~\cite{lee2019set} principles. It uses task-optimized dual multiset functions that maintain permutation invariance, crucial for hypergraph learning. The update rules in AllSet are:
\begin{align}
    Z_{e,:}^{(t+1), v} &= f_{\mathcal{V}\rightarrow \mathcal{E}} (V_{e\backslash v, X^{(t)}}; Z_{e, :}^{(t), v}, X_{v, :}^{(t)}), \\
    X_{v,:}^{(t+1)} &= f_{\mathcal{E}\rightarrow \mathcal{V}} (E_{v, Z^{(t+1), v}}; X_{v, :}^{(t)}).\label{eq:allset}
\end{align}
Here, $f_{\mathcal{V}\rightarrow \mathcal{E}}$ and $f_{\mathcal{E}\rightarrow \mathcal{V}}$ are the key multiset functions mapping node and hyperedge features. For example, $f_{\mathcal{V}\rightarrow \mathcal{E}}(S) = \text{MLP}\left(\sum_{s\in S} \text{MLP}(s)\right)$ is used in AllDeepSets. The notation $V_{e, X}$ and $E_{v,Z}$ represent multisets of node and hyperedge features, respectively. The AllSet approach updates nodes and hyperedges in the hypergraph by leveraging their features in conjunction with those of adjacent hyperedges or nodes, enabling a rich representation of the hypergraph structure. The method could differentiate node $v$ from its multiset, allowing for sophisticated feature aggregation.

\section{Methods}
We introduce the SE3Set model to leverage hypergraph neural networks for capturing complex molecular interactions, integrating both 2D chemical and 3D spatial structures (Sec.~\ref{sec:fragmentation}). It builds upon the AllSet framework~\cite{chien2021you} and the Equiformer~\cite{liao2022equiformer}. Upcoming sections will delve into the specifics of molecular fragmentation and the SE3Set architecture.

\begin{figure*}[ht]
\begin{center}
\centerline{\includegraphics[width=1.0\columnwidth]{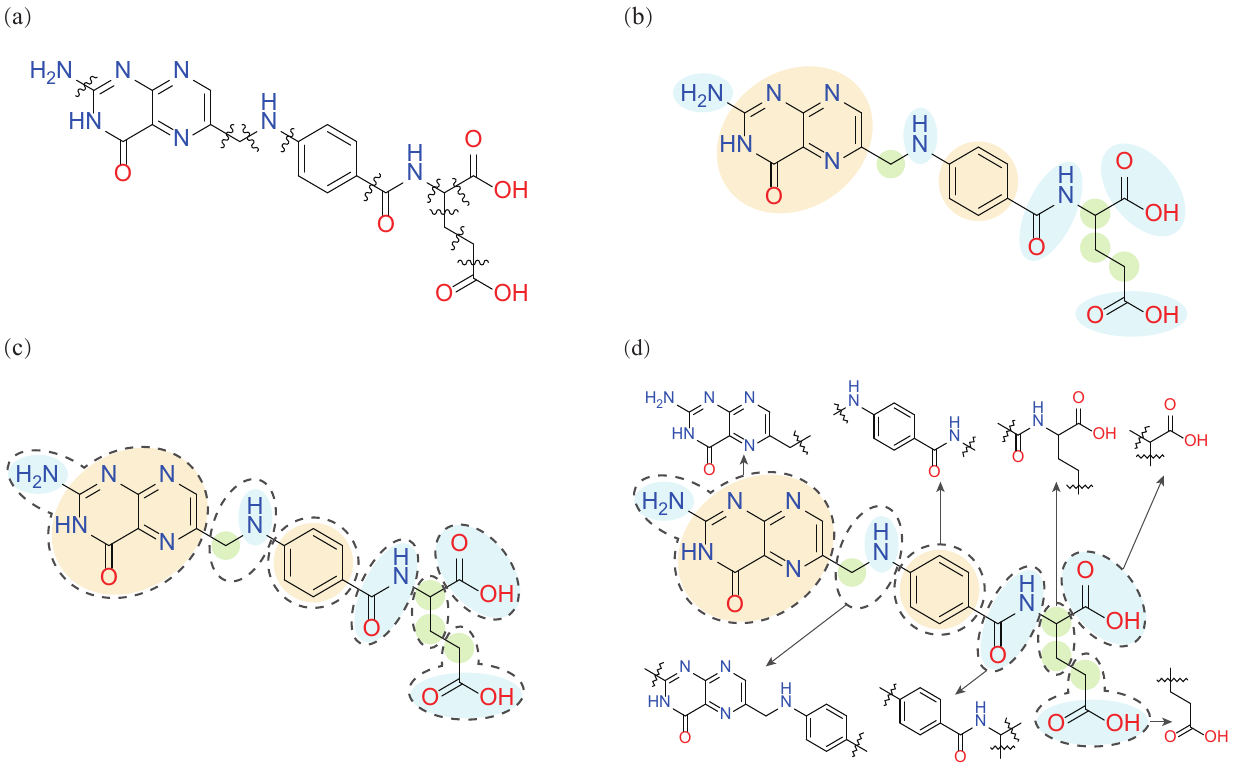}}
\caption{Folic acid fragmentation illustrated with CID 135398658 from PubChem. (a) Preprocessing to identify cleavable bonds for fragmentation. (b) Initial fragments formed using BFS, color-coded by functional groups (blue), rings (orange), and single atoms (green). (c) Fragments merged to satisfy atom count criteria, detailed in~\ref{appendix:fragment_rules}. (d) Expansion of fragments shown with directional arrows.}
\label{fig:fragmentation}
\end{center}
\vspace{-1.0cm}
\end{figure*}

\subsection{Fragmentation algorithm}\label{sec:fragmentation}
To harness the power of hypergraph neural networks for molecular representations, we need to map molecules onto hypergraph structures through a refined fragmentation algorithm. Our strategy intertwines molecular topology and spatial geometry to create hyperedges that capture groups of atoms, reflecting their functional and spatial characteristics. In crafting this fragmentation approach for hypergraph-based molecular representation, the methodology must adhere to a set of fundamental principles:

\begin{enumerate}
    \item The design should merge topological chemistry with 3D structural data into a unified hypergraph representation, ensuring hyperedges accurately embody the molecule's chemical and spatial properties.
    \item Controlling fragment size is vital for the SE3Set model to balance capturing meaningful interactions and computational efficiency. Optimal fragment sizes are key for model performance and learning capabilities.
    \item The fragmentation could only selectively break single bonds and must maintain functional groups and ring integrity to preserving key chemical information critical for the molecule's properties and behavior.
    \item Fragment overlap is essential to maintain functional group effects on local charge distribution and to ensure hyperedge interaction within the SE3Set model for improved molecular learning.
\end{enumerate}

Before delving deeper into the specifics of our fragmentation method, it's important to establish a foundational understanding through key definitions and concepts, 
\begin{definition}
    The bond order represents the multiplicity or the number of shared electron pairs that constitute a covalent bond between two atoms.
\end{definition}
The bond order matrix $\mathbf{B}$ is an $N \times N$ representation of bond strength between atoms in a molecule, with higher bond order values indicating stronger bonds. This symmetric matrix ($B_{ij} = B_{ji}$) is crucial for studying molecular structure and reactivity, capturing bond nuances including delocalized and resonance bonds in computational chemistry.

Our fragmentation algorithm improves molecular representations by combining bond order, functional groups, and substructures, including SMARTS-identified smaller rings and merged adjacent groups. Overcoming the drawbacks of non-overlapping fragmentation, it uses 3D spatial data and allows overlaps, preserving local effects for precise charge distribution and enhancing hypergraph neural network learning of molecular interactions.

\begin{definition}
    A molecular fragment, denoted as $\mathfrak{F}$, is defined as a specific subset of atoms within a molecule, characterized by being a cohesive assembly of predefined substructures linked in a sequential concatenation.
\end{definition}

Our fragmentation method meticulously dissects a given molecule into meaningful subsets of atoms, and this process unfolds through four steps (corresponding to the pipeline in Fig.~\ref{fig:fragmentation}):

\begin{enumerate}
    \item Pre-processing by analyzing the molecule's bond order matrix to mask high-order bonds and those within functional groups or rings, and merging adjacent functional groups for a streamlined structural representation.
    \item Core substructures are delineated from the remaining bonds using a Breadth-First Search algorithm, establishing the basic units of the molecular framework.
    \item These substructures are then aggregated into larger molecular fragments according to predefined rules that maintain a minimum atom count within each fragment. This step could be optional.
    \item To enhance fragment connectivity, we expand each by incorporating adjacent groups, using interaction strength metrics based on interatomic distances to guide this process. Here we set the cutoff value denoted as $c_w$ of an interaction strength metrics to intercept the extended fragment.\label{enmt:expand_frag_explicit}
\end{enumerate}

Furthermore, step~\ref{enmt:expand_frag_explicit} leads to a substantial computational overhead for hypergraph neural networks when processing larger molecular systems. To enhance the efficiency of our model for such expansive molecular systems, we introduce a revised strategy for step~\ref{enmt:expand_frag_explicit}: 

\begin{enumerate}
     \item[4*] \label{enmt:expand_frag_implicit} For each atom $i$, identify the neighboring atoms $\mathcal{N}_i$ that fall within a specified radial cutoff $r_c$. A fragment $\mathfrak{F}$ is considered to be adjacent to atom $i$ if there is an overlap of at least one atom between $\mathfrak{F}$ and $\mathcal{N}_i$. For ease of reference, the set of fragments adjacent to atom $i$ is represented as $\mathcal{N}^\mathcal{F}_i$, which implies that $\mathcal{N}^\mathcal{F}_i = \{\mathfrak{F}| \mathfrak{F} \cap \mathcal{N}_i \neq \varnothing\}$.
\end{enumerate}

We designate the application of step~\ref{enmt:expand_frag_explicit} as an~\emph{explicit overlap} and the application of step \hyperref[enmt:expand_frag_implicit]{4*} as an~\emph{implicit overlap}. These approaches introduce nuanced variations in the mathematical expressions of our model, as reflected in the Eq.~\ref{eq:V2E_aggr_explicit}, Eq.~\ref{eq:V2E_aggr_implicit}, and Eq.~\ref{eq:E2V_aggr}. For the detailed step-by-step methodology, please refer to Appendix~\ref{appendix:fragment_steps}.

\begin{figure*}[htp]
\begin{center}
\centerline{\includegraphics[width=1.0\textwidth]{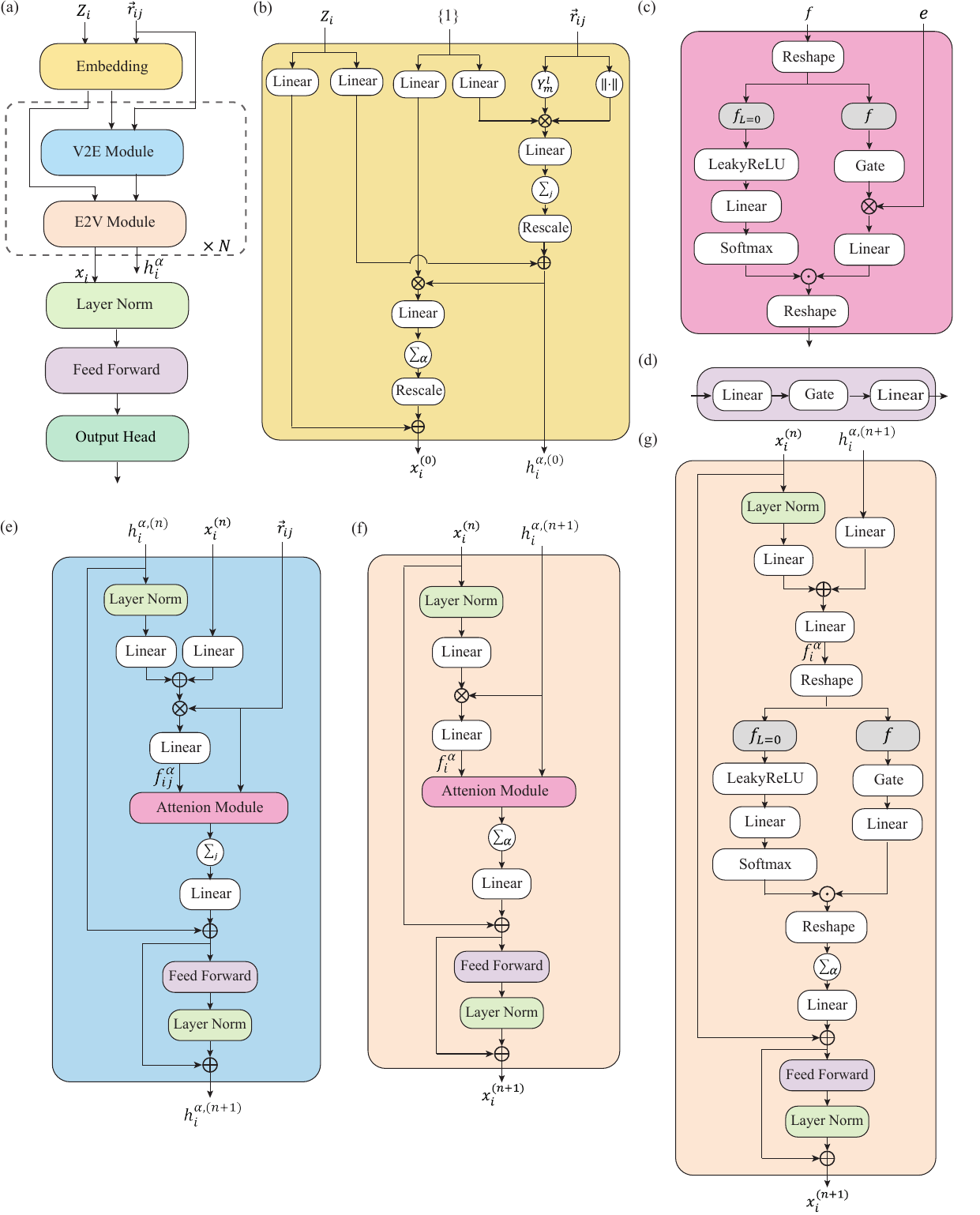}} 
\caption{Overall architecture of SE3Set. (a) SE3Set begins with node and hyperedge embeddings, cycles through V2E and E2V attention modules for iterative updates, and concludes with normalization and a feed-forward block for output. (b) Embedding. Atomic numbers and position vectors are transformed into initial embeddings for nodes and hyperedges. (c) Attention Block. Merges feature sets with positional or hyperedge data for feature processing. (d) Feed-Forward Block. Enhances feature sets through a streamlined network. (e) V2E Module. Utilizes node features and their relative positions to update hyperedge features. (f) E2V Module. Employs hyperedge features to refresh node features, using tensor products (left) or summation (right) for updates. Symbols $\otimes$, $\oplus$, and $\odot$ in figures denote depth-wise tensor product, summation, and Hadamard multiplication, respectively. $h_i^{\alpha}$ represents hyperedge features, $x_i$ is for node features, superscript $n$ indicates the number of updates, and $\vec{r}_{ij}$ is the relative position vector between nodes $i$ and $j$.}
\label{fig:se3set}
\end{center}
\end{figure*}

\subsection{SE3Set}\label{sec:se3set}
Building upon our aforementioned fragmentation algorithm, we now turn to outline the architecture of SE3Set. SE3Set model, influenced by AllSet~\cite{chien2021you} and built on the Equiformer~\cite{liao2022equiformer}, incorporates 3D spatial equivariance in our hypergraph neural network, improving capture of many-body interactions for precise molecular structure representation. SE3Set consists of an embedding layer, attention blocks, and an output head, as shown in Fig.~\ref{fig:se3set} (a).

\subsubsection{Embedding}
As depicted in Fig.~\ref{fig:se3set} (b), the embedding block generates detailed node and hyperedge features reflecting molecular structures. Node features blend intrinsic properties with degree embeddings from connected hyperedges, while hyperedge features aggregate node embeddings. They're mapped onto $l$-order SH functions for SE3 equivariance and updated separately. Hyperedges capture nodes' positional relationships, assigning a distinct feature $h^{\alpha}_i$ to each node $i$ in hyperedge $\mathfrak{F}_{\alpha}$, reinforcing structural fidelity. Nodes $x_i$ integrate hyperedge information, harmonizing uniqueness with interconnections. Attention mechanisms then refine node and hyperedge interactions for accurate molecular and structural representation.

\subsubsection{Equivariant hypergraph attention blocks}
As presented in Fig.~\ref{fig:se3set} (c)-(f), the attention mechanism comprises two essential components: the Vertex-to-Edge (V2E) and Edge-to-Vertex (E2V) attention blocks, based on the AllSet framework~\cite{chien2021you}. The V2E block refines hyperedge features, while the E2V block updates node features, both operating with an equivariant hypergraph attention mechanism. To improve training and enable deeper network structures, we incorporate normalization layers and residual connections to prevent gradient issues. The attention module's output passes through a feed forward block (Fig.~\ref{fig:se3set} (d)), enhancing representation complexity. Node and hyperedge features maintain equivariance to molecular geometry, preserving data symmetries and the integrity of representations, thus boosting the model's expressiveness in capturing complex structural interactions. (The concepts of irreducible representations and tensor products can be referenced in the Appendix~\ref{appendix: group-theory}.)

\paragraph{V2E attention} The SE3Set model uses geometrically invariant attention weights $a_{ij}$, derived from $l=0$ irreps acting as scalars under geometric transformations. These weights are computed from scalar features $f_{ij, l=0}$ using an MLP with LeakyReLU activation and softmax normalization, reflecting node relationships within the hypergraph. Node and hyperedge features undergo non-linear transformations represented by tensor products of irreps with quantum number $l$. The features combine through direct tensor products (DTP), yielding non-linear values $v_{ij}$ (Fig.~\ref{fig:se3set} (c)). Hyperedge features are updated by aggregating features from connected nodes, utilizing SH and radial basis functions on hyperedge features. The model calculates initial features $f_{ij}^{\alpha}$ and V2E attention weights $a_{ij}^{\alpha}$ via MLPs, with non-linear values $v_{ij}^{\alpha}$ emerging from similar transformations.
\begin{align}
    t_{ij}^\alpha &= (\text{Linear}(x_i) + \text{Linear}(x_j)) \\
    f_{ij}^\alpha &= \text{Linear}(t_{ij}^\alpha \otimes^{\text{DTP}}_{w(\Vert\vec{r}_{ij}\Vert)} \text{SH}(\vec{r}_{ij}))\\
    a_{ij}^\alpha &= \text{Softmax}_j(a^\top \text{LeakyReLU}(f_{ij, l=0}^\alpha))\\
    v_{ij}^\alpha &= \text{Linear}(\text{Gate}(f_{ij}^\alpha) \otimes^{\text{DTP}}_{w(\Vert\vec{r}_{ij}\Vert)} \text{SH}(\vec{r}_{ij}))
\end{align}

Ultimately, the SE3Set model updates hyperedge features $h_i^k$ by accumulating the weighted features of nodes within the same hyperedge and applying a linear transformation to the aggregated information. For the \emph{explicit overlap} fragmentation method, 
\begin{equation}
    \Delta h_i^{\alpha} = \text{Linear}\left(\sum_{j: n_i\in \mathfrak{F}_\alpha \land n_j\in\mathfrak{F}_\alpha} a_{ij}^\alpha v_{ij}^\alpha\right) \label{eq:V2E_aggr_explicit}
\end{equation}
where $n_i$ denotes the node with index $i$ and $\mathfrak{F}_\alpha$ denotes the fragment with index $\alpha$ as each fragment could be considered as a hyperedge in the hypergraph. Due to the frequent occurrence of a high number of explicitly overlapping atoms, this scenario commonly results in increased computational complexity. Consequently, when adopting the \emph{implicit overlap} approach, we may opt for an equation of the form:
\begin{equation}
    \Delta h_i^{\alpha} = \text{Linear}\left(\sum_{j: j\in \mathfrak{F}_\alpha \land \mathfrak{F}_\alpha \in \mathcal{N}^{\mathcal{F}}_i} a_{ij}^\alpha v_{ij}^\alpha\right) \label{eq:V2E_aggr_implicit}
\end{equation}
where $\mathcal{N}^{\mathcal{F}}_i$ is delineated in step \hyperref[enmt:expand_frag_implicit]{4*} of the \emph{implicit overlap} method. This characteristic renders it a more computationally efficient scheme for Vertex-to-Edge (V2E) attention mechanisms. The detailed architecture of V2E attention block is shown in Fig.~\ref{fig:se3set} (e).

\paragraph{E2V attention} Following the V2E attention module, the E2V attention module (Fig.~\ref{fig:se3set} (f)) updates node features by transforming them with a tensor product of the updated hyperedge feature, followed by a linear layer. Attention weights are then calculated using softmax-applied, LeakyReLU-activated features, ensuring node features are refined after hyperedge updates.
\begin{align}
\label{eq:E2V_tp_begin}
    f_{i}^\alpha &= \text{Linear}((\text{Linear}(x_i) \otimes^{\text{DTP}} h_i^\alpha)\\
    a_{i}^\alpha &= \text{Softmax}_\alpha(a^\top \text{LeakyReLU}(f_{i, l=0}^\alpha))
\end{align}
These attention weights direct the synthesis of information, culminating in the calculated value:
\begin{equation}
    v_{i}^\alpha = \text{Linear}(\text{Gate}(f_{i}^\alpha) \otimes^{\text{DTP}} h_i^\alpha).
\label{eq:E2V_tp_end}
\end{equation}

Furthermore, we propose an alternative method for constructing the E2V attention block as shown in Fig.~\ref{fig:se3set} (g).
\begin{align}
\label{eq:E2V_sum_begin}
    f_{i}^\alpha &= \text{Linear}(\text{Linear}(x_i) + \text{Linear}(h_i^\alpha))\\
    a_{i}^\alpha &= \text{Softmax}_\alpha(a^\top \text{LeakyReLU}(f_{i, l=0}^\alpha))\\
    v_{i}^\alpha &= \text{Linear}(\text{Gate}(f_{i}^\alpha))\label{eq:E2V_sum_end}
\end{align}
However, practical experiments reveal that the previous method yields superior results, with detailed findings presented in Sec.~\ref{sec:ablation}.

Then the node aggregates all the hyperedge features corresponding to itself to update the node feature, 
\begin{align}\label{eq:E2V_aggr}
    \text{Explicit overlap: }\Delta x_i = \text{Linear}\left(\sum_{\alpha: i \in \mathfrak{F}\alpha} a_{i}^\alpha v_{i}^\alpha\right), \\ 
    \text{Implicit overlap: }\Delta x_i = \text{Linear}\left(\sum_{\alpha: \mathfrak{F}_\alpha \in \mathcal{N}^{\mathcal{F}}_i} a_{i}^\alpha v_{i}^\alpha\right)
\end{align}

\subsubsection{Output head}
The SE3Set model employs node features to generate predictions, using a feed-forward network to transform these features into the target label's irreps dimension. A summation strategy aggregates node features into a single hypergraph-level representation, which is then processed by a linear layer to output the model's final predictions.

\section{Results}
We tested our equivariant hypergraph neural network on QM9~\cite{ruddigkeit2012enumeration, ramakrishnan2014quantum}, MD17~\cite{chmiela2017machine} (see Appendix~\ref{appendix:md17}), and MD22~\cite{chmiela2023accurate} to assess its molecular representation learning. QM9 and MD17 gauge small molecule property prediction, while MD22 evaluates larger systems with complex many-body interactions~\cite{wang2023efficiently}. An ablation study was also conducted to pinpoint the contributions of fragmentation and architecture to our method's performance, offering insights into the network's efficacy and areas for enhancement.

\begin{table*}[hbtp]
\caption{A comparative analysis was performed to assess the Mean Absolute Errors (MAEs) on the QM9 dataset when training SE3Set on a configuration comprising 110,000 training samples and 1,000 validation samples. Bolding shows the best model and underlining shows the second best model and the underlining tilde shows third best model.}
\label{tab:qm9}
\begin{center}
\begin{small}
\begin{sc}
\resizebox{\linewidth}{!}{
\begin{tabular}{lcccccccccccc}
\toprule
~ & unit & SchNet & DimeNet++ & PaiNN & SphereNet & ComENet & ET & Allegro & ViSNet & QuinNet & Equiformer & SE3Set \\
\midrule
$\mu$    & $D$ & 0.033 & 0.030 & 0.012 & 0.026 & 0.0245 & \uline{0.011} & - & \textbf{0.010} & 0.771 & \uline{0.011} & \uline{0.011}\\
$\alpha$ & $a_0^3$ & 0.235 & \uline{0.044} & \uwave{0.045} & 0.046 & 0.0452 & 0.059 & - & \textbf{0.041} & 0.047 & 0.046 & \uwave{0.045}\\
HOMO     & $\rm{meV}$ & 41 & 25 & 20 & 23 & 23 & 20 & - & \uwave{17.3} & 20.4 & \textbf{15} & \textbf{15}\\
LUMO     & $\rm{meV}$ & 34 & 20 & 28 & 18 & 20 & 18 & - & \uwave{14.8} & 17.6 & \uline{14} & \textbf{13}\\
gap      & $\rm{meV}$ & 63 & 33 & 46 & 32 & 32 & 36 & - & 31.7 & \textbf{28.2} & \uwave{30} & \uline{29}\\
$R^2$    & $a_0^2$ & 0.073 & 0.331 & \uwave{0.066} & 0.292 & 0.259 & \uline{0.033} & - & \textbf{0.030} & 0.194 & 0.251 & 0.197\\
ZPVE     & $\rm{meV}$ & 1.70 & \uwave{1.21} & 1.28 & \textbf{1.12} & \uline{1.20} & 1.84 & - & 1.56 & 1.26 & 1.26 & 1.40\\
$U_0$    & $\rm{meV}$ & 14 & 6 & 5.85 & 6 & 6.59 & 6.15 & \uline{4.7} & \textbf{4.23} & 7.6 & 6.59 & \uwave{5.74}\\
$U$      & $\rm{meV}$ & 19 & 6 & 5.83 & 7 & 6.82 & 6.38 & \uline{4.4} & \textbf{4.25} & 8.4 & 6.74 & \uwave{5.69}\\
$H$      & $\rm{meV}$ & 14 & 7 & 5.98 & 6 & 6.86 & 6.16 & \textbf{4.4} & \uline{4.52} & 7.8 & 6.63 & \uwave{5.70}\\
$G$      & $\rm{meV}$ & 14 & 8 & 7.35 & 8 & 7.98 & 7.62 & \textbf{5.7} & \uline{5.86} & 8.5 & 7.63 & \uwave{6.63}\\
$C_v$    & $\rm{\frac{kcal}{mol\cdot K}}$ & 0.033 & \uline{0.023} & 0.024 & \textbf{0.021} & 0.024 & 0.026 & - & \uline{0.023} & 0.024 & \uline{0.023} & 0.025\\
\bottomrule
\end{tabular}}
\end{sc}
\end{small}
\end{center}
\end{table*}

\subsection{QM9}
The QM9 dataset~\cite{ruddigkeit2012enumeration, ramakrishnan2014quantum} consists of 134k small organic molecules calculated at the B3LYP/6-31G(2df, p) level. SE3Set, after training on 110k QM9 molecules and validation on 10k, achieves low mean absolute errors (MAEs) in 12 tasks, performing on par with leading models, as detailed in Table~\ref{tab:qm9}. In small molecular systems, higher-order many-body interactions are less pronounced, and as a result, SE3Set does not significantly outperform other state-of-the-art (SOTA) models.

\begin{table*}[hbtp]
\caption{A comparison of Mean Absolute Errors (MAEs) across various benchmarked models. SE3Set is trained on the five molecules of MD22 dataset with specific number of training/validation. Bolding shows the best model and underlining shows the second best model. The improvements column shows the improvement of our model over the previous SOTA model in percentage terms. The MAEs reflect the precision of energy predictions in units of kcal/mol and forces in units of kcal/(mol$\cdot$\AA). The results of TorchMD-Net, Allegro, and Equiformer are extracted from Ref.~\cite{li2024longshortrange}}
\label{tab:md22}
\begin{center}
\begin{small}
\begin{sc}
\centering
\resizebox{\linewidth}{!}{
\begin{tabular}{lllccccccccccc}
\toprule
Molecule & \# Train/Val & & sGDML & TorchMD-NET & Allegro & MACE & Equiformer & ViSNet & QuinNet & Equiformer-LSRM & ViSNet-LSRM & SE3Set & Improvements\\
\midrule
\multirow{2}{*}{Ac-Ala3-NHMe} & \multirow{2}{*}{5500/500} & Energy & 0.3902 & 0.1121 & 0.1019 & \underline{0.0620} & 0.0828 &  0.0796 & 0.084 & 0.0780 & 0.0654 & \textbf{0.0499} & 19.5\%\\
~ & ~ & Force & 0.7968 & 0.1879 & 0.1068 & 0.0876 & 0.0804 &  0.0972 & \underline{0.0681} & 0.0877 & 0.0902 & \textbf{0.0545} & 20.0\%\\
\midrule
\multirow{2}{*}{DHA} & \multirow{2}{*}{7500/500} & Energy & 1.3117 & 0.1205 & 0.1153 & 0.1317 & 0.1788 &  0.1526 & 0.12 & 0.0878 & \underline{0.0873} & \textbf{0.0826} & 5.4\%\\
~ & ~ & Force & 0.7474 & 0.1209 & 0.0732 & 0.0646 & \underline{0.0506} &  0.0668 & 0.0515 & 0.0534 & 0.0598 & \textbf{0.0360} & 28.9\%\\
\midrule
\multirow{2}{*}{Stachyose} & \multirow{2}{*}{7500/500} & Energy & 4.0497 & 0.1393 & 0.2485 & 0.1244 & 0.1404 & 0.1283 & 0.23 & 0.1252 & \underline{0.1055} & \textbf{0.0762} & 27.8\%\\
~ & ~ & Force & 0.6744 & 0.1921 & 0.0971 & 0.0876 & 0.0635 & 0.0869 & \underline{0.0543} & 0.0632 & 0.0767 & \textbf{0.0424} & 21.9\%\\
\midrule
\multirow{2}{*}{AT-AT} & \multirow{2}{*}{2500/500} & Energy & 0.7235 & 0.1120 & 0.1428 & 0.1093 & 0.1309 & 0.1688 & 0.14 & 0.1007 & \underline{0.0772} & \textbf{0.0585} & 24.2\%\\
~ & ~ & Force & 0.6911 & 0.2036 & 0.0952 & 0.0992 & 0.0960 &  0.1070 & \underline{0.0687} & 0.0811 & 0.0781 & \textbf{0.0556} & 19.1\%\\
\midrule
\multirow{2}{*}{AT-AT-CG-CG} & \multirow{2}{*}{1500/500} & Energy & 1.3885 & 0.2072 & 0.3933 & 0.1578 & 0.1510 & 0.1995 & 0.38 & 0.1335 & \underline{0.1135} & \textbf{0.1002} & 11.7\%\\
~ & ~ & Force & 0.7028 & 0.3259 & 0.1280 & 0.1153 & 0.1252 & 0.1563 & 0.1273 & 0.1065 & \underline{0.1063} & \textbf{0.0825} & 22.4\%\\
\bottomrule
\end{tabular}}
\end{sc}
\end{small}
\end{center}
\vspace{-0.4cm}
\end{table*}

\subsection{MD22}\label{sec:md22}
Recognizing the prominence of higher-order many-body interactions in larger molecules~\cite{wang2023efficiently}, SE3Set was tested on the comprehensive MD22 dataset~\cite{chmiela2023accurate}. This dataset spans four classes of biomolecules and supramolecules, from a 42-atom peptide to a 370-atom nanotube, with high-resolution sampling at 400-500~K using the PBE+MBD~\cite{perdew1996generalized, tkatchenko2012accurate} framework for energy and force computations. Our fragmentation method, which maintains functional groups and rings, selectively excludes structures like the Buckyball catcher and Double-walled nanotube from MD22, thus concentrating on the other five molecular types. As Table~\ref{tab:md22} shows, SE3Set outperforms other SOTA models in these cases, reducing MAEs by an average of 20\%, underscoring its exceptional ability to capture molecular intricacies. Moreover, our results indicate that incorporating higher-order many-body interactions is crucial for representing the non-local features of larger molecules within the MD22 dataset.

\begin{wrapfigure}{r}{8.5cm} \vspace{-1.1cm}
\includegraphics[width=0.6\columnwidth]{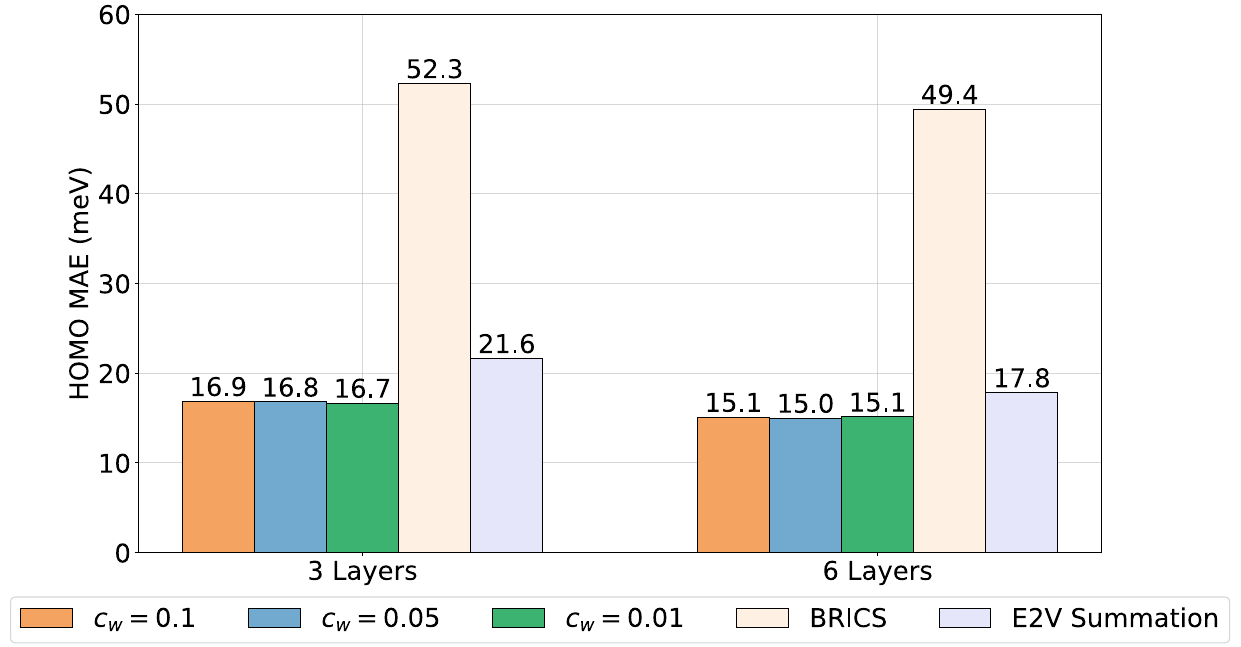}
\centering
\caption{Ablation studies on the QM9 dataset's HOMO task (units: $\rm{meV}$). The variable $c_w$ represents the threshold for expansion in the fourth step of fragmentation, guided by the fragment bond order defined in Eq.~\ref{eq:bo_lendvay}. The term BRICS denotes another fragmentation method implemented in RDKits. Additionally, the E2V summation refers to the architectural framework specified from Eq.~\ref{eq:E2V_sum_begin} to Eq.~\ref{eq:E2V_sum_end}.}
\label{fig:ablation}
\vspace{-0.2cm}
\end{wrapfigure}

\subsection{Ablation studies}\label{sec:ablation}
To better understand SE3Set, we conduct ablation studies focusing on fragmentation and model architecture. We explore how different fragmentation techniques affects SE3Set's training and compare with the non-overlapping BRICS~\cite{degen2008art, greg_landrum_2020_3732262} strategy. As Fig.~\ref{fig:ablation} indicates, tests on QM9's homo energy task showed SE3Set's robustness to $c_w$ variations in fragmentation method. Our method surpasses BRICS demonstrates the importance of hyperedge interaction. Furthermore, we performed ablation studies on the model architecture. Among two design variants in the E2V attention section, the one using tensor product interactions between nodes and hyperedges proved superior, emphasizing the value of our tensor product-based mechanism and architecture design in enhancing molecular property predictions. Additionally, a 6-layer SE3Set model outperformed its 3-layer counterpart.

\section{Conclusion}
In conclusion, this study demonstrate the efficacy of SE3Set, a cutting-edge hypergraph neural network architecture, in the realm of molecular representation learning. By meticulously crafting a fragmentation method that coalesces two-dimensional chemical knowledge with three-dimensional spatial information, we establish a robust foundation for constructing hypergraphs that faithfully capture the complex nature of molecular structures. The SE3Set architecture, drawing inspiration from the AllSet framework and the Equiformer, adeptly processing these hypergraphs and preserving the essential invariances and symmetries. SE3Set demonstrates performance comparable to SOTA models in small molecular systems and significantly outperforms SOTA models in large molecular systems where higher-order many-body interactions are pronounced. The results of our research affirm the potential of SE3Set to model high-order many-body interactions, providing a powerful tool for molecular representation.

\bibliographystyle{unsrt}
\bibliography{ref}

\begin{thebibliography}{10}

\bibitem{mathews2012molecular}
Jonathan~P Mathews and Alan~L Chaffee.
\newblock The molecular representations of coal--a review.
\newblock {\em Fuel}, 96:1--14, 2012.

\bibitem{david2020molecular}
Laurianne David, Amol Thakkar, Roc{\'\i}o Mercado, and Ola Engkvist.
\newblock Molecular representations in {AI}-driven drug discovery: a review and practical guide.
\newblock {\em J. Cheminform.}, 12(1):1--22, 2020.

\bibitem{wigh2022review}
Daniel~S Wigh, Jonathan~M Goodman, and Alexei~A Lapkin.
\newblock A review of molecular representation in the age of machine learning.
\newblock {\em Wiley Interdiscip. Rev. Comput. Mol. Sci.}, 12(5):e1603, 2022.

\bibitem{fourches2010trust}
Denis Fourches, Eugene Muratov, and Alexander Tropsha.
\newblock Trust, but verify: on the importance of chemical structure curation in cheminformatics and qsar modeling research.
\newblock {\em J. Chem. Inf. Model.}, 50(7):1189, 2010.

\bibitem{zhou2020graph}
Jie Zhou, Ganqu Cui, Shengding Hu, Zhengyan Zhang, Cheng Yang, Zhiyuan Liu, Lifeng Wang, Changcheng Li, and Maosong Sun.
\newblock Graph neural networks: A review of methods and applications.
\newblock {\em AI open}, 1:57--81, 2020.

\bibitem{wu2020comprehensive}
Zonghan Wu, Shirui Pan, Fengwen Chen, Guodong Long, Chengqi Zhang, and S~Yu Philip.
\newblock A comprehensive survey on graph neural networks.
\newblock {\em IEEE Trans. Neural. Netw. Learn. Syst.}, 32(1):4--24, 2020.

\bibitem{do2019graph}
Kien Do, Truyen Tran, and Svetha Venkatesh.
\newblock Graph transformation policy network for chemical reaction prediction.
\newblock In {\em Proceedings of the 25th ACM SIGKDD international conference on knowledge discovery \& data mining}, pages 750--760, 2019.

\bibitem{xiong2021graph}
Jiacheng Xiong, Zhaoping Xiong, Kaixian Chen, Hualiang Jiang, and Mingyue Zheng.
\newblock Graph neural networks for automated de novo drug design.
\newblock {\em Drug Discov. Today}, 26(6):1382--1393, 2021.

\bibitem{reiser2022graph}
Patrick Reiser, Marlen Neubert, Andr{\'e} Eberhard, Luca Torresi, Chen Zhou, Chen Shao, Houssam Metni, Clint van Hoesel, Henrik Schopmans, Timo Sommer, et~al.
\newblock Graph neural networks for materials science and chemistry.
\newblock {\em Commun. Mater.}, 3(1):93, 2022.

\bibitem{gilmer2017neural}
Justin Gilmer, Samuel~S Schoenholz, Patrick~F Riley, Oriol Vinyals, and George~E Dahl.
\newblock Neural message passing for quantum chemistry.
\newblock In {\em International conference on machine learning}, pages 1263--1272. PMLR, 2017.

\bibitem{godwin2021simple}
Jonathan Godwin, Michael Schaarschmidt, Alexander~L Gaunt, Alvaro Sanchez-Gonzalez, Yulia Rubanova, Petar Veli{\v{c}}kovi{\'c}, James Kirkpatrick, and Peter Battaglia.
\newblock Simple {GNN} regularisation for 3{D} molecular property prediction and beyond.
\newblock In {\em International Conference on Learning Representations}, 2021.

\bibitem{rusch2023survey}
T~Konstantin Rusch, Michael~M Bronstein, and Siddhartha Mishra.
\newblock A survey on oversmoothing in graph neural networks.
\newblock {\em Preprint at \url{http://arxiv.org/abs/2303.10993}}, 2023.

\bibitem{gasteiger2019directional}
Johannes Gasteiger, Janek Gro{\ss}, and Stephan G{\"u}nnemann.
\newblock Directional message passing for molecular graphs.
\newblock In {\em International Conference on Learning Representations}, 2019.

\bibitem{schutt2021equivariant}
Kristof Sch{\"u}tt, Oliver Unke, and Michael Gastegger.
\newblock Equivariant message passing for the prediction of tensorial properties and molecular spectra.
\newblock In {\em International Conference on Machine Learning}, pages 9377--9388. PMLR, 2021.

\bibitem{batzner2022}
Simon Batzner, Albert Musaelian, Lixin Sun, Mario Geiger, Jonathan~P Mailoa, Mordechai Kornbluth, Nicola Molinari, Tess~E Smidt, and Boris Kozinsky.
\newblock E (3)-equivariant graph neural networks for data-efficient and accurate interatomic potentials.
\newblock {\em Nat. Commun.}, 13(1):2453, 2022.

\bibitem{gasteiger2020fast}
Johannes Gasteiger, Shankari Giri, Johannes~T Margraf, and Stephan G{\"u}nnemann.
\newblock Fast and uncertainty-aware directional message passing for non-equilibrium molecules.
\newblock {\em Preprint at \url{http://arxiv.org/abs/2011.14115}}, 2020.

\bibitem{gasteiger2021gemnet}
Johannes Gasteiger, Florian Becker, and Stephan G{\"u}nnemann.
\newblock Gemnet: Universal directional graph neural networks for molecules.
\newblock {\em Advances in Neural Information Processing Systems}, 34:6790--6802, 2021.

\bibitem{tholke2021equivariant}
Philipp Th{\"o}lke and Gianni De~Fabritiis.
\newblock Equivariant transformers for neural network based molecular potentials.
\newblock In {\em International Conference on Learning Representations}, 2021.

\bibitem{batatia2022mace}
Ilyes Batatia, David~P Kovacs, Gregor Simm, Christoph Ortner, and G{\'a}bor Cs{\'a}nyi.
\newblock {MACE}: Higher order equivariant message passing neural networks for fast and accurate force fields.
\newblock {\em Advances in Neural Information Processing Systems}, 35:11423--11436, 2022.

\bibitem{musaelian2023learning}
Albert Musaelian, Simon Batzner, Anders Johansson, Lixin Sun, Cameron~J Owen, Mordechai Kornbluth, and Boris Kozinsky.
\newblock Learning local equivariant representations for large-scale atomistic dynamics.
\newblock {\em Nat. Commun.}, 14(1):579, 2023.

\bibitem{wang2024enhancing}
Yusong Wang, Tong Wang, Shaoning Li, Xinheng He, Mingyu Li, Zun Wang, Nanning Zheng, Bin Shao, and Tie-Yan Liu.
\newblock Enhancing geometric representations for molecules with equivariant vector-scalar interactive message passing.
\newblock {\em Nat. Commun.}, 15(1):313, 2024.

\bibitem{wang2023efficiently}
Zun Wang, Guoqing Liu, Yichi Zhou, Tong Wang, and Bin Shao.
\newblock Efficiently incorporating quintuple interactions into geometric deep learning force fields.
\newblock In {\em Thirty-seventh Conference on Neural Information Processing Systems}, 2023.

\bibitem{thomas2018tensor}
Nathaniel Thomas, Tess Smidt, Steven Kearnes, Lusann Yang, Li~Li, Kai Kohlhoff, and Patrick Riley.
\newblock Tensor field networks: Rotation-and translation-equivariant neural networks for 3d point clouds.
\newblock {\em Preprint at \url{http://arxiv.org/abs/1802.08219}}, 2018.

\bibitem{anderson2019cormorant}
Brandon Anderson, Truong~Son Hy, and Risi Kondor.
\newblock Cormorant: Covariant molecular neural networks.
\newblock {\em Advances in neural information processing systems}, 32, 2019.

\bibitem{fuchs2020se}
Fabian Fuchs, Daniel Worrall, Volker Fischer, and Max Welling.
\newblock S{E} (3)-transformers: 3{D} {R}oto-translation equivariant attention networks.
\newblock {\em Advances in Neural Information Processing Systems}, 33:1970--1981, 2020.

\bibitem{liao2022equiformer}
Yi-Lun Liao and Tess Smidt.
\newblock Equiformer: Equivariant graph attention transformer for 3{D} atomistic graphs.
\newblock In {\em The Eleventh International Conference on Learning Representations}, 2022.

\bibitem{liao2023equiformerv2}
Yi-Lun Liao, Brandon Wood, Abhishek Das, and Tess Smidt.
\newblock Equiformer{V}2: Improved equivariant transformer for scaling to higher-degree representations.
\newblock {\em Preprint at \url{http://arxiv.org/abs/2306.12059}}, 2023.

\bibitem{schutt2017schnet}
Kristof Sch{\"u}tt, Pieter-Jan Kindermans, Huziel~Enoc Sauceda~Felix, Stefan Chmiela, Alexandre Tkatchenko, and Klaus-Robert M{\"u}ller.
\newblock Schnet: A continuous-filter convolutional neural network for modeling quantum interactions.
\newblock {\em Advances in neural information processing systems}, 30, 2017.

\bibitem{schutt2018schnet}
P-J Kindermans and K-R M{\"u}ller.
\newblock Schnet--a deep learning architecture for molecules and materials.
\newblock {\em J. Chem. Phys.}, 148(24), 2018.

\bibitem{coors2018spherenet}
Benjamin Coors, Alexandru~Paul Condurache, and Andreas Geiger.
\newblock Sphere{N}et: Learning spherical representations for detection and classification in omnidirectional images.
\newblock In {\em Proceedings of the European conference on computer vision (ECCV)}, pages 518--533, 2018.

\bibitem{wang2022comenet}
Limei Wang, Yi~Liu, Yuchao Lin, Haoran Liu, and Shuiwang Ji.
\newblock Com{EN}et: Towards complete and efficient message passing for 3{D} molecular graphs.
\newblock In Alice~H. Oh, Alekh Agarwal, Danielle Belgrave, and Kyunghyun Cho, editors, {\em Advances in Neural Information Processing Systems}, 2022.

\bibitem{agarwal2005beyond}
Sameer Agarwal, Jongwoo Lim, Lihi Zelnik-Manor, Pietro Perona, David Kriegman, and Serge Belongie.
\newblock Beyond pairwise clustering.
\newblock In {\em 2005 IEEE Computer Society Conference on Computer Vision and Pattern Recognition (CVPR'05)}, volume~2, pages 838--845. IEEE, 2005.

\bibitem{zhou2006learning}
Dengyong Zhou, Jiayuan Huang, and Bernhard Sch{\"o}lkopf.
\newblock Learning with hypergraphs: Clustering, classification, and embedding.
\newblock {\em Advances in neural information processing systems}, 19, 2006.

\bibitem{li2013z}
Guoyin Li, Liqun Qi, and Gaohang Yu.
\newblock The {Z}-eigenvalues of a symmetric tensor and its application to spectral hypergraph theory.
\newblock {\em Numer. Linear Algebra Appl.}, 20(6):1001--1029, 2013.

\bibitem{pearson2014spectral}
Kelly~J Pearson and Tan Zhang.
\newblock On spectral hypergraph theory of the adjacency tensor.
\newblock {\em Graphs Combin.}, 30:1233--1248, 2014.

\bibitem{benson2017spacey}
Austin~R Benson, David~F Gleich, and Lek-Heng Lim.
\newblock The spacey random walk: A stochastic process for higher-order data.
\newblock {\em SIAM Rev.}, 59(2):321--345, 2017.

\bibitem{chien2021landing}
Eli Chien, Pan Li, and Olgica Milenkovic.
\newblock Landing probabilities of random walks for seed-set expansion in hypergraphs.
\newblock In {\em 2021 IEEE Information Theory Workshop (ITW)}, pages 1--6. IEEE, 2021.

\bibitem{tudisco2021nonlinear}
Francesco Tudisco, Austin~R Benson, and Konstantin Prokopchik.
\newblock Nonlinear higher-order label spreading.
\newblock In {\em Proceedings of the Web Conference 2021}, pages 2402--2413, 2021.

\bibitem{kim2021transformers}
Jinwoo Kim, Saeyoon Oh, and Seunghoon Hong.
\newblock Transformers generalize deepsets and can be extended to graphs \& hypergraphs.
\newblock {\em Advances in Neural Information Processing Systems}, 34:28016--28028, 2021.

\bibitem{kim2022equivariant}
Jinwoo Kim, Saeyoon Oh, Sungjun Cho, and Seunghoon Hong.
\newblock Equivariant hypergraph neural networks.
\newblock In {\em European Conference on Computer Vision}, pages 86--103. Springer, 2022.

\bibitem{cui2023hyper}
Shicheng Cui, Qianmu Li, Deqiang Li, Zhichao Lian, Jun Hou, et~al.
\newblock Hyper-mol: Molecular representation learning via fingerprint-based hypergraph.
\newblock {\em Comput. Intell. Neurosci.}, 2023, 2023.

\bibitem{tavakoli2022rxn}
Mohammadamin Tavakoli, Alexander Shmakov, Francesco Ceccarelli, and Pierre Baldi.
\newblock Rxn hypergraph: a hypergraph attention model for chemical reaction representation.
\newblock {\em Preprint at \url{http://arxiv.org/abs/2201.01196}}, 2022.

\bibitem{kajino2019molecular}
Hiroshi Kajino.
\newblock Molecular hypergraph grammar with its application to molecular optimization.
\newblock In {\em International Conference on Machine Learning}, pages 3183--3191. PMLR, 2019.

\bibitem{nachmani2020molecule}
Eliya Nachmani and Lior Wolf.
\newblock Molecule property prediction and classification with graph hypernetworks.
\newblock {\em Preprint at \url{http://arxiv.org/abs/2002.00240}}, 2020.

\bibitem{chen2021hypergraph}
Fangying Chen, Junyoung Park, and Jinkyoo Park.
\newblock A hypergraph convolutional neural network for molecular properties prediction using functional group.
\newblock {\em Preprint at \url{http://arxiv.org/abs/2106.01028}}, 2021.

\bibitem{chen2023molecular}
Junwu Chen and Philippe Schwaller.
\newblock Molecular hypergraph neural networks.
\newblock {\em Preprint at \url{http://arxiv.org/abs/2312.13136}}, 2023.

\bibitem{gordon2012fragmentation}
Mark~S Gordon, Dmitri~G Fedorov, Spencer~R Pruitt, and Lyudmila~V Slipchenko.
\newblock Fragmentation methods: A route to accurate calculations on large systems.
\newblock {\em Chem. Rev.}, 112(1):632--672, 2012.

\bibitem{collins2015energy}
Michael~A Collins and Ryan~PA Bettens.
\newblock Energy-based molecular fragmentation methods.
\newblock {\em Chem. Rev.}, 115(12):5607--5642, 2015.

\bibitem{du2021hypergraph}
Boxin Du, Changhe Yuan, Robert Barton, Tal Neiman, and Hanghang Tong.
\newblock Hypergraph pre-training with graph neural networks.
\newblock {\em Preprint at \url{http://arxiv.org/abs/2105.10862}}, 2021.

\bibitem{kim2022contrastive}
Seojin Kim, Jaehyun Nam, Junsu Kim, Hankook Lee, Sungsoo Ahn, and Jinwoo Shin.
\newblock Contrastive learning of molecular representation with fragmented views.
\newblock 2022.

\bibitem{luong2023fragment}
Kha-Dinh Luong and Ambuj Singh.
\newblock Fragment-based pretraining and finetuning on molecular graphs.
\newblock {\em Preprint at \url{http://arxiv.org/abs/2310.03274}}, 2023.

\bibitem{chien2021you}
Eli Chien, Chao Pan, Jianhao Peng, and Olgica Milenkovic.
\newblock You are {A}ll{S}et: A multiset function framework for hypergraph neural networks.
\newblock In {\em International Conference on Learning Representations}, 2021.

\bibitem{zaheer2017deep}
Manzil Zaheer, Satwik Kottur, Siamak Ravanbakhsh, Barnabas Poczos, Russ~R Salakhutdinov, and Alexander~J Smola.
\newblock Deep sets.
\newblock {\em Advances in neural information processing systems}, 30, 2017.

\bibitem{lee2019set}
Juho Lee, Yoonho Lee, Jungtaek Kim, Adam Kosiorek, Seungjin Choi, and Yee~Whye Teh.
\newblock Set transformer: A framework for attention-based permutation-invariant neural networks.
\newblock In {\em International conference on machine learning}, pages 3744--3753. PMLR, 2019.

\bibitem{ruddigkeit2012enumeration}
Lars Ruddigkeit, Ruud Van~Deursen, Lorenz~C Blum, and Jean-Louis Reymond.
\newblock Enumeration of 166 billion organic small molecules in the chemical universe database {GDB}-17.
\newblock {\em J. Chem. Inf. Model.}, 52(11):2864--2875, 2012.

\bibitem{ramakrishnan2014quantum}
Raghunathan Ramakrishnan, Pavlo~O Dral, Matthias Rupp, and O~Anatole Von~Lilienfeld.
\newblock Quantum chemistry structures and properties of 134 kilo molecules.
\newblock {\em Sci. Data}, 1(1):1--7, 2014.

\bibitem{chmiela2017machine}
Stefan Chmiela, Alexandre Tkatchenko, Huziel~E Sauceda, Igor Poltavsky, Kristof~T Sch{\"u}tt, and Klaus-Robert M{\"u}ller.
\newblock Machine learning of accurate energy-conserving molecular force fields.
\newblock {\em Sci. Adv.}, 3(5):e1603015, 2017.

\bibitem{chmiela2023accurate}
Stefan Chmiela, Valentin Vassilev-Galindo, Oliver~T Unke, Adil Kabylda, Huziel~E Sauceda, Alexandre Tkatchenko, and Klaus-Robert M{\"u}ller.
\newblock Accurate global machine learning force fields for molecules with hundreds of atoms.
\newblock {\em Science Advances}, 9(2):eadf0873, 2023.

\bibitem{li2024longshortrange}
Yunyang Li, Yusong Wang, Lin Huang, Han Yang, Xinran Wei, Jia Zhang, Tong Wang, Zun Wang, Bin Shao, and Tie-Yan Liu.
\newblock Long-short-range message-passing: A physics-informed framework to capture non-local interaction for scalable molecular dynamics simulation.
\newblock In {\em The Twelfth International Conference on Learning Representations}, 2024.

\bibitem{perdew1996generalized}
John~P Perdew, Kieron Burke, and Matthias Ernzerhof.
\newblock Generalized gradient approximation made simple.
\newblock {\em Phys. Rev. Lett.}, 77(18):3865, 1996.

\bibitem{tkatchenko2012accurate}
Alexandre Tkatchenko, Robert~A DiStasio~Jr, Roberto Car, and Matthias Scheffler.
\newblock Accurate and efficient method for many-body van der {W}aals interactions.
\newblock {\em Phys. Rev. Lett.}, 108(23):236402, 2012.

\bibitem{degen2008art}
J{\"o}rg Degen, Christof Wegscheid-Gerlach, Andrea Zaliani, and Matthias Rarey.
\newblock On the art of compiling and using'drug-like'chemical fragment spaces.
\newblock {\em ChemMedChem: Chemistry Enabling Drug Discovery}, 3(10):1503--1507, 2008.

\bibitem{greg_landrum_2020_3732262}
Greg Landrum, Paolo Tosco, Brian Kelley, sriniker, gedeck, NadineSchneider, Riccardo Vianello, Ric, Andrew Dalke, Brian Cole, AlexanderSavelyev, Matt Swain, Samo Turk, Dan N, Alain Vaucher, Eisuke Kawashima, Maciej Wójcikowski, Daniel Probst, guillaume godin, David Cosgrove, Axel Pahl, JP, Francois Berenger, strets123, JLVarjo, Noel O'Boyle, Patrick Fuller, Jan~Holst Jensen, Gianluca Sforna, and DoliathGavid.
\newblock rdkit/rdkit: 2020\_03\_1 (q1 2020) release, March 2020.

\bibitem{Lendvay2000OnTC}
Gy{\"o}rgy Lendvay.
\newblock On the correlation of bond order and bond length.
\newblock {\em J. Mol. Struct.}, 501:389--393, 2000.

\bibitem{Bridgeman2006BondOB}
Adam~J. Bridgeman and Christopher~J Empson.
\newblock Bond orders between molecular fragments.
\newblock {\em Chemistry}, 12 8:2252--62, 2006.

\bibitem{mobley2018escaping}
David~L Mobley, Caitlin~C Bannan, Andrea Rizzi, Christopher~I Bayly, John~D Chodera, Victoria~T Lim, Nathan~M Lim, Kyle~A Beauchamp, David~R Slochower, Michael~R Shirts, et~al.
\newblock Escaping atom types in force fields using direct chemical perception.
\newblock {\em J. Chem. Theory Comput.}, 14(11):6076--6092, 2018.

\bibitem{jeff_wagner_2024_10593535}
Jeff Wagner, Matt Thompson, David~L. Mobley, John Chodera, Caitlin Bannan, Andrea Rizzi, trevorgokey, David~L. Dotson, Josh~A. Mitchell, jaimergp, Camila, Pavan Behara, Christopher Bayly, JoshHorton, Iván Pulido, Lily Wang, Victoria Lim, Sukanya Sasmal, SimonBoothroyd, Andrew Dalke, Daniel Smith, Brent Westbrook, Josh Horton, Lee-Ping Wang, Richard Gowers, Ziyuan Zhao, Connor Davel, and Yutong Zhao.
\newblock {openforcefield/openff-toolkit: 0.15.1. Testing updates}, January 2024.

\bibitem{fey2019fast}
Matthias Fey and Jan~Eric Lenssen.
\newblock Fast graph representation learning with pytorch geometric.
\newblock {\em Preprint at \url{http://arxiv.org/abs/1903.02428}}, 2019.

\end{thebibliography}



\newpage

\appendix

\section*{Appendix}

\section{Details of fragmentation steps}\label{appendix:fragment_steps}
Based on the design principles in Sec.~\ref{sec:fragmentation}, the detailed step-by-step methodology of \emph{explicit overlap} fragmentation method is shown as follows, 

\begin{enumerate}
    \item The pre-processing step begins by analyzing the given molecule through its bond order matrix, denoted as $B$. Identify and mask bonds that are part of functional groups or rings, as well as those with a bond order of $B_{ij} \geqslant 2$. Functional groups are then identified using predefined SMARTS patterns for accurate matching. To achieve a more generalized representation of functional groups, topologically adjacent functional groups are merged into a single entity. This aggregation allows to focus on specific subfunctional groups that are of particular interest, simplifying the complexity of the molecular structure for subsequent analysis.
    \item Following the masking of selected bonds, the Breadth-First Search (BFS) algorithm is employed to reconstruct the substructures, denoted as $\{\mathfrak{S}\}$, from the remaining unmasked bonds. These groups represent the core structural units of the molecule as discussed at the outset of this section.
    \item Consolidate the previously identified groups $\{\mathfrak{S}\}$ into larger molecular fragments, applying a set of predefined rules to guide the merging process. These rules are meticulously designed to ensure that each resulting fragment, now denoted as $\{\mathfrak{F}\}$, contains at least a minimum specified number of atoms. For a comprehensive understanding of the merging criteria, one can refer to the detailed rules outlined in~\ref{appendix:fragment_rules}.
    \item Extend each fragment $\{\mathfrak{F}\}$ by incorporating adjacent groups from $\{\mathfrak{S}\}$ to enrich the connectivity between molecular fragments, thus intentionally creating regions of overlap among the fragments. This expansion is controlled by a cutoff threshold, denoted as $c_w$, which is typically a function based on interatomic distances. the fragment bond order~\cite{Lendvay2000OnTC, Bridgeman2006BondOB}, symbolized by $W_{fs}$, is used to quantitatively assess the interaction strength between a fragment $\mathfrak{F}_i$ and an adjacent substructure $\mathfrak{S}_j$. This method reflects the interaction strength based on the proximity of atoms in different fragments, expressed by the following equation:
    \begin{equation}
        W_{fs} = \sum_{i\in \mathfrak{F}_f, j\in \mathfrak{S}_s}\exp \left(-\frac{(d_{ij} - d^e_{ij})\cdot d^e_{ij}}{(0.25~\text{\AA})^2}\right), \label{eq:bo_lendvay}
    \end{equation}
    where $d_{ij}$ represents the interatomic distance between atoms $i$ and $j$, and $d^e_{ij}$ stands for the equilibrium distance typically expected for such a bond. This equation is utilized to determine which substructures should be included in the expansion of a fragment, based on the strength of their interactions as governed by the distance function. Additionally, in alignment with Pauling's concept of "chemist's bond order"~\cite{Lendvay2000OnTC}, an alternative method is introduced to calculate the bond order using a single exponential function, 
    \begin{equation}
        W_{fs} = \sum_{i\in \mathfrak{F}_f, j\in \mathfrak{S}_s}\exp \left(-(d_{ij} - d^e_{ij})\right), \label{eq:bo_exp}
    \end{equation}
    where $W_{fs}$ encapsulates the bond order between atoms belonging to a fragment $\mathfrak{F}_f$ and a substructure $\mathfrak{S}_s$. In this context, $d_{ij}$ signifies the actual measured distance between atom $i$ and atom $j$. The term $d^e_{ij}$ refers to the theoretical equilibrium covalent bond length, which is estimated by summing the empirical covalent radii of the two atoms involved, given by:
    \begin{equation}
        d_{ij}^e = r_{z_i} + r_{z_j}, 
    \end{equation}
    where $r_{z_i}$ is the empirical covalent radius of an atom with atomic number $z_i$. This function provides a simplified yet effective representation of bond order, allowing us to gauge the bonding interactions within the molecular structure with respect to the proximity of the atoms.
\end{enumerate}

The \emph{implicit overlap} fragmentation method only change the step \ref{enmt:expand_frag_explicit}, the details of the changed fourth step has been spelled out in \ref{enmt:expand_frag_implicit}.

\section{Functional groups SMARTS}
In the initial phase of our fragmentation approach, we identify functional groups using the SMARTS pattern matching language. In Table~\ref{tab:functional-groups}, we present the complete list of SMARTS patterns utilized, which have been expanded upon from the default set found within the Open Force Field toolkit~\cite{mobley2018escaping, jeff_wagner_2024_10593535} (accessible at: \url{https://github.com/openforcefield/openff-fragmenter/blob/main/openff/fragmenter/data/default-functional-groups.json}).

\begin{table}[htp]
\caption{SMARTS patterns for functional groups employed in the preprocessing stage of fragmentation.}
\label{tab:functional-groups}
\vskip 0.15in
\begin{center}
\begin{small}
\begin{sc}
\begin{tabular}{lcccr}
\toprule
Functional Groups Name & SMARTS  \\
\midrule
hydrazine & \text{[NX3:1][NX3:2]} \\
hydrazone & \text{[NX3:1][NX2:2]} \\
nitric oxide & \text{[N:1]-[O:2]} \\
amide & \text{[\#7:1][\#6:2](=[\#8:3]), [NX3:1][CX3:2](=[OX1:3])[NX3:4]}\\
amide negative ion & [\#7:1][\#6:2](-[O-:3])\\
aldehyde & \text{[CX3H1:1](=[O:2])[\#6:3]}\\
sulfoxide & \text{[\#16X3:1]=[OX1:2], [\#16X3+:1][OX1-:2]}\\
sulfonyl & \text{[\#16X4:1](=[OX1:2])=[OX1:3]}\\
sulfinic acid & \text{[\#16X3:1](=[OX1:2])[OX2H,OX1H0-:3]}\\
sulfonic acid & \text{[\#16X4:1](=[OX1:2])(=[OX1:3])[OX2H,OX1H0-:4]}\\
sulfinamide & \text{[\#16X4:1](=[OX1:2])(=[OX1:3])([NX3R0:4])}\\
phosphine oxide & \text{[PX4:1](=[OX1:2])([\#6:3])([\#6:4])([\#6:5])}\\
phosphonate & \text{[P:1](=[OX1:2])([OX2H,OX1-:3])([OX2H,OX1-:4])}\\
phosphate & \text{[PX4:1](=[OX1:2])([\#8:3])([\#8:4])([\#8:5])}\\
carboxylic acid & \text{[CX3:1](=[O:2])[OX1H0-,OX2H1:3]}\\
nitro & \text{[NX3+:1](=[O:2])[O-:3], [NX3:1](=[O:2])=[O:3]}\\
ester & \text{[CX3:1](=[O:2])[OX2H0:3]}\\
tri-halide & \text{[\#6:1]([F,Cl,I,Br:2])([F,Cl,I,Br:3])([F,Cl,I,Br:4])}\\
hydroxyl & \text{[\#8:1]-[\#1:2]}\\
\bottomrule
\end{tabular}
\end{sc}
\end{small}
\end{center}
\vskip -0.1in
\end{table}

\section{Merge process of fragmentation}\label{appendix:fragment_rules}
During the third step of our fragmentation method, we introduce a strategy to enlarge substructures, ensuring that each initial fragment contains at least $n_{\min}$ atoms, with $n_{\min}$ being a predefined integer. To maintain permutation invariance for a molecule, we incorporate weights, $W_{fs}$, to guide the sequence of merging. The process is outlined in the pseudocode (Algorithm~\ref{alg:frag_merge}). The calculation of $W$ is based on either Eq.~\ref{eq:bo_lendvay} or Eq.~\ref{eq:bo_exp}. By considering the sum of bond orders to other groups, we assess each group's centrality. The groups are then ordered first by the number of atoms they contain, followed by the summation of their bond orders, ensuring that the fragmentation merge process is permutation invariant when following this specified sequence. The algorithm then assists smaller groups in merging with others to achieve a size of at least $n_{\min}$ atoms. Initially, we consider topologically adjacent groups with the fewest atoms. If a target group lacks topological neighbors, we proceed to merge based on the bond order from $W$. We introduce a threshold $c_{is}$ that allows a group to remain isolated if it is significantly distant from others. It should be noted that isolated groups may not meet the minimum atom number requirement; however, they could be further expanded in the subsequent fragmentation step, depending on the chosen thresholds for $c_{is}$ and $c_w$ (refer to Sec.~\ref{sec:fragmentation}). Overall, this algorithm ensures a permutation invariant merging process.

\begin{algorithm}[tb]
   \caption{Pseudo code of fragmentation merge step.}
   \label{alg:frag_merge}
\begin{algorithmic}
   \STATE {\bfseries Input:} groups $\{\mathfrak{G}\}$, minimum atoms number $n_{\min}$, maximum atoms number $n_{\max}$, Topological bond order matrix $B$, isolated threshold $c_{is}$
   \STATE $m=\vert\{\mathfrak{G}\}\vert$
   \STATE Isolate groups $\{\mathfrak{G^I}\} = \{\}$
   \STATE Calculate fragmentation bond order matrix $W_{\mathfrak{G}_i\mathfrak{G}_j}$.
   \STATE Sort $\{\mathfrak{G}\}$ in descending order based on the following attributes: number of atoms, $\sum_{\mathfrak{G}', \mathfrak{G}'\neq \mathfrak{G}} W_{\mathfrak{G}\mathfrak{G}'}$.
   \REPEAT 
   \STATE Pop last fragment as $\mathfrak{G}_k$ from $\{\mathfrak{G}\}$
   \FOR{$i=m-1$ {\bfseries to} $1$}
   \STATE $a = \text{MAX\_INT}, merge\_idx = -1$ 
   \IF{ {\bfseries any} $B_{ij} \geqslant 1, i\in \mathfrak{G}_i  j\in \mathfrak{G}_k$ {\bfseries and} $\vert\mathfrak{G}_i\vert < a$ {\bfseries and} $a + \vert\mathfrak{G}_i\vert \leqslant n_{\max}$}
   \STATE $a = \vert\mathfrak{G}_i\vert, merge\_idx = i$
   \ENDIF
   \ENDFOR

   \IF{$merge\_idx == -1$}
   \FOR{$i=m-1$ {\bfseries to} $1$}
   \IF{ {\bfseries any} $W_{\mathfrak{G}_i\mathfrak{G}_k} \geqslant c_{is}$ {\bfseries and} $\vert\mathfrak{G}_i\vert < a$}
   \STATE $a = \vert\mathfrak{G}_i\vert, merge\_idx = -1$
   \ENDIF
   \ENDFOR
   \ENDIF

   \IF{$merge\_idx \neq -1$}
   \STATE Merge $\mathfrak{G}_k$ to $\mathfrak{G}_{merge\_idx}$
   \STATE Resort $\{\mathfrak{G}\}$ by the same priority and update $W$.
   \ELSE 
   \STATE Add $\{\mathfrak{G}_k\}$ to $\{\mathfrak{G^I}\}$
   \ENDIF
   
   \UNTIL{$\vert\{\mathfrak{G}_k\}\vert \geqslant n_{\min}$}
   \STATE $\{\mathfrak{F}\} = \{\mathfrak{G}\}\cup\{\mathfrak{G^I}\}$\\

\end{algorithmic}
\end{algorithm}

\section{Distribution of fragmentation dataset}
Different parameters used in the fragmentation process can lead to a variety of hyperedges, which in turn result in distinct hypergraphs utilized for training our model. To illustrate the variances attributed to different fragmentation parameters or methods (such as BRICS implemented in RDKit~\cite{degen2008art, greg_landrum_2020_3732262}), we use the QM9 dataset~\cite{ruddigkeit2012enumeration, ramakrishnan2014quantum} to demonstrate how the data distributions attached to hypergraphs may change.

The impact of adjusting fragmentation parameters on the composition of hyperedges can be observed in Fig.~\ref{fig:frag_dis}. Altering the expansion threshold $c_w$ within a certain range has a minimal effect on fragment expansion. However, when utilizing the Lendvay bond order (Eq.~\ref{eq:bo_lendvay}), fragments tend to comprise fewer atoms compared to when using the Exponential bond order (Eq.~\ref{eq:bo_exp}). This difference is likely due to the more gradual decline in the exponential function, which results in a greater cumulative contribution to the weights $W_{fs}$.

Our ablation study (Sec.~\ref{sec:ablation}) also includes a comparison with the BRICS fragmentation method. Fragments generated by the BRICS method are observed to contain significantly fewer atoms since this approach does not create overlapping regions between different fragments.

\section{Concepts of irreps features and tensor product}\label{appendix: group-theory}
\paragraph{Irreps features} The SE3Set model utilizes the special orthogonal group SO(3) to capture three-dimensional rotational symmetries in molecular structures. This approach is similar to Equiformer~\cite{liao2022equiformer, liao2023equiformerv2}.It employs irreducible representations (irreps) of SO(3), parameterized by an integer $l$, which correspond to spherical harmonics (SH) functions $Y_l^m$. These functions imbue feature vectors with rotational information, ensuring the model's equivariance to rotations and enabling consistent geometric property analysis. This approach is key to the model's ability to accurately represent and predict molecular and other rotationally invariant systems.

\paragraph{Tensor product} To boost the model's expressive power, we consider interactions between irrep features of different angular momenta $l$ through the tensor product, which merges two irreps $l_1$ and $l_2$ into a new irrep with angular momentum $l_3$. This is achieved using Clebsch-Gordan coefficients in an expansion weighted by $w_{m_1,m_2}$.
\begin{equation}
\begin{split}
    f^{l_3}_{m_3} & =(f^{l_1}_{m_1} \otimes f^{l_2}_{m_2}) _{m_3} \\
    &= \sum_{m_1, m_2} w_{m_1,m_2}C_{l_1,m_1,~ l_2, m_2}^{l_3, m_3} f^{l_1}_{m_1} f^{l_2}_{m_2}.
\end{split}
\end{equation}
To reduce complexity, a depth-wise tensor product $\otimes^{\text{DTP}}$ is adopted from the Equiformer~\cite{liao2022equiformer, liao2023equiformerv2}, utilizing internal weights to streamline computations. Input-dependent tensor product weights are denoted as $\otimes^{\text{DTP}}_w$, ensuring computational efficiency while preserving equivariance for feature interactions.

\begin{table*}[hbtp]
\caption{A comparison of Mean Absolute Errors (MAEs) across various benchmarked models. SE3Set is trained on the MD17 dataset with a configuration of 950 training samples and 50 validation samples. Bolding shows the best model and underlining shows the second best model and the underlining tilde shows third best model. The MAEs reflect the precision of energy predictions in units of kcal/mol and forces in units of kcal/(mol$\cdot$\AA). }
\label{tab:md17}
\begin{center}
\begin{small}
\begin{sc}
\centering
\resizebox{\linewidth}{!}{
\begin{tabular}{llcccccccccccc}
\toprule
~ & ~ & SchNet & DimeNet & PaiNN & ET & GemNet & NequIP ($l$=3) & ViSNet & QuinNet & Equiformer & SE3Set\\
\midrule
\multirow{2}{*}{Aspirin} & Energy & 0.37 & 0.204 & 0.167 & 0.123 & - & 0.131 &  \textbf{0.116} & \uline{0.119} & \uwave{0.122} & 0.130\\
~ & Force & 1.35 & 0.499 & 0.338 & 0.253 & 0.217 & 0.184 &  0.155 & \textbf{0.145} & \uline{0.152} & \uwave{0.153}\\
\midrule
\multirow{2}{*}{Ethanol} & Energy & 0.08 & 0.064 & 0.064 & 0.052 & - & \uline{0.051} & \uline{0.051} & \textbf{0.050} & \uline{0.051} & 0.054\\
~ & Force & 0.39 & 0.230 & 0.224 & 0.109 & 0.085 & 0.071 & \textbf{0.060} & \textbf{0.060} & 0.067 & \uwave{0.062}\\
\midrule
\multirow{2}{*}{Malonaldehyde} & Energy & 0.13 & 0.104 & 0.091 & 0.077 & - & 0.076 & \uline{0.075} & 0.078 & \textbf{0.074} & \textbf{0.074} \\
~ & Force & 0.66 & 0.383 & 0.319 & 0.169 & 0.155 & 0.129& \uline{0.100} & \textbf{0.097} & 0.125 & \uwave{0.103}\\
\midrule
\multirow{2}{*}{Naphthalene} & Energy & 0.16 & 0.122 & 0.116 & 0.085 & - & 0.113 & \textbf{0.085} & \uwave{0.101} & \textbf{0.085} & 0.113\\
~ & Force & 0.58 & 0.215 & 0.077 & 0.061 & 0.051 & \textbf{0.039} & \textbf{0.039} & \textbf{0.039} & 0.046 & \textbf{0.039}\\
\midrule
\multirow{2}{*}{Salicylic acid} & Energy & 0.20 & 0.134 & 0.116 & 0.093 & - & 0.106 & \textbf{0.092} & \uwave{0.101} & \uline{0.099} & 0.108\\
~ & Force & 0.85 & 0.374 & 0.195 & 0.129 & 0.125 & \uwave{0.090} & \uline{0.084} & \textbf{0.080} & \uwave{0.090} & \uwave{0.090}\\
\midrule
\multirow{2}{*}{Toluene} & Energy & 0.12 & 0.102 & 0.095 & 0.074 & - & 0.092& \textbf{0.074} & \uline{0.080} & \uwave{0.085} & 0.093\\
~ & Force & 0.57 & 0.216 & 0.094 & 0.067 & 0.060 & \uwave{0.046} & \textbf{0.039} & \textbf{0.039} & 0.048 & \uwave{0.046}\\
\midrule
\multirow{2}{*}{Uracil} & Energy & 0.14 & 0.115 & 0.106 & \uline{0.095} & - & 0.104 & \textbf{0.095} & \uwave{0.096} & 0.099 & 0.103\\
~ & Force & 0.56 & 0.301 & 0.139 & 0.095 & 0.097 & 0.076 & \textbf{0.062} & \textbf{0.062} & 0.076 & \uwave{0.067}\\
\bottomrule
\end{tabular}}
\end{sc}
\end{small}
\end{center}
\end{table*}

\section{Results of MD17}\label{appendix:md17}
The MD17 dataset~\cite{chmiela2017machine} features a wide variety of molecular configurations simulated at 500 K, with high-resolution trajectories and labeled with energies and forces from the PBE+vdW-TS method~\cite{perdew1996generalized, tkatchenko2012accurate}. SE3Set's performance on this dataset is shown in Table~\ref{tab:md17}. SE3Set outperforms Equiformer in accuracy, highlighting its refined force calculation capabilities. In small molecular systems, higher-order many-body interactions are less pronounced, and as a result, SE3Set does not significantly outperform other state-of-the-art (SOTA) models.

\section{Training details}\label{sec:training}
This section outlines the training specifics, encompassing the fragmentation parameters, SE3Set hyperparameters, and certain implementation nuances utilized in our experimental setup.

Our dataset construction is founded on PyTorch Geometric~\cite{fey2019fast} augmented with our fragmentation process (Sec.~\ref{sec:fragmentation}). Due to inconsistencies in molecular topology identified through RDKit's sanitization routine~\cite{greg_landrum_2020_3732262}, 1,403 data points were excised from the original dataset. We designated 110,000 data points for the training set and 10,000 for the validation set, selected at random. 

We used an \emph{explicit overlap} scheme on the QM9 and MD17 datasets because of their relatively small molecular systems. We implemented two distinct schemes for calculating fragment bond orders, following either Eq.~\ref{eq:bo_lendvay} or Eq.~\ref{eq:bo_exp}. The parameters for fragmentation are detailed below. Given that the MD17 molecules are relatively small, the merge step in the fragmentation process was not actually utilized. However, we present the fragmentation parameters here for the sake of completeness.

Besides, we adopt an \emph{implicit overlap} scheme on the MD22 dataset to reduce computational resource consumption. The details of cutoff $r_c$ introduced in \hyperref[enmt:expand_frag_implicit]{4*} can be found in Table~\ref{tab:hyperparams-frag-md22rc}.

On QM9 and MD17 dataset, our model was trained using a single Tesla V100 GPU with 32GB of memory, except for the 6-layer model employing the exponential bond order on QM9 dataset, which was trained on two Tesla V100 GPUs with 32GB each. For MD22 dataset, our model was trained on a single Tesla A100 GPU with 80GB of memory.

We selected $l = 2$ for our irreducible representations (irreps) feature, which includes both node and hypergraph features. For the radial basis function (RBF), we utilized Gaussian basis functions or Bessel basis functions for the QM9 dataset~\cite{ruddigkeit2012enumeration, ramakrishnan2014quantum} and exponential basis functions for the MD17 dataset and MD22 dataset~\cite{chmiela2017machine}. Details could be found in Table~\ref{tab:hyperparams-frag} and Table~\ref{tab:hyperparams-model}.

\begin{figure}[H]
\begin{center}
\centerline{\includegraphics[width=1\columnwidth]{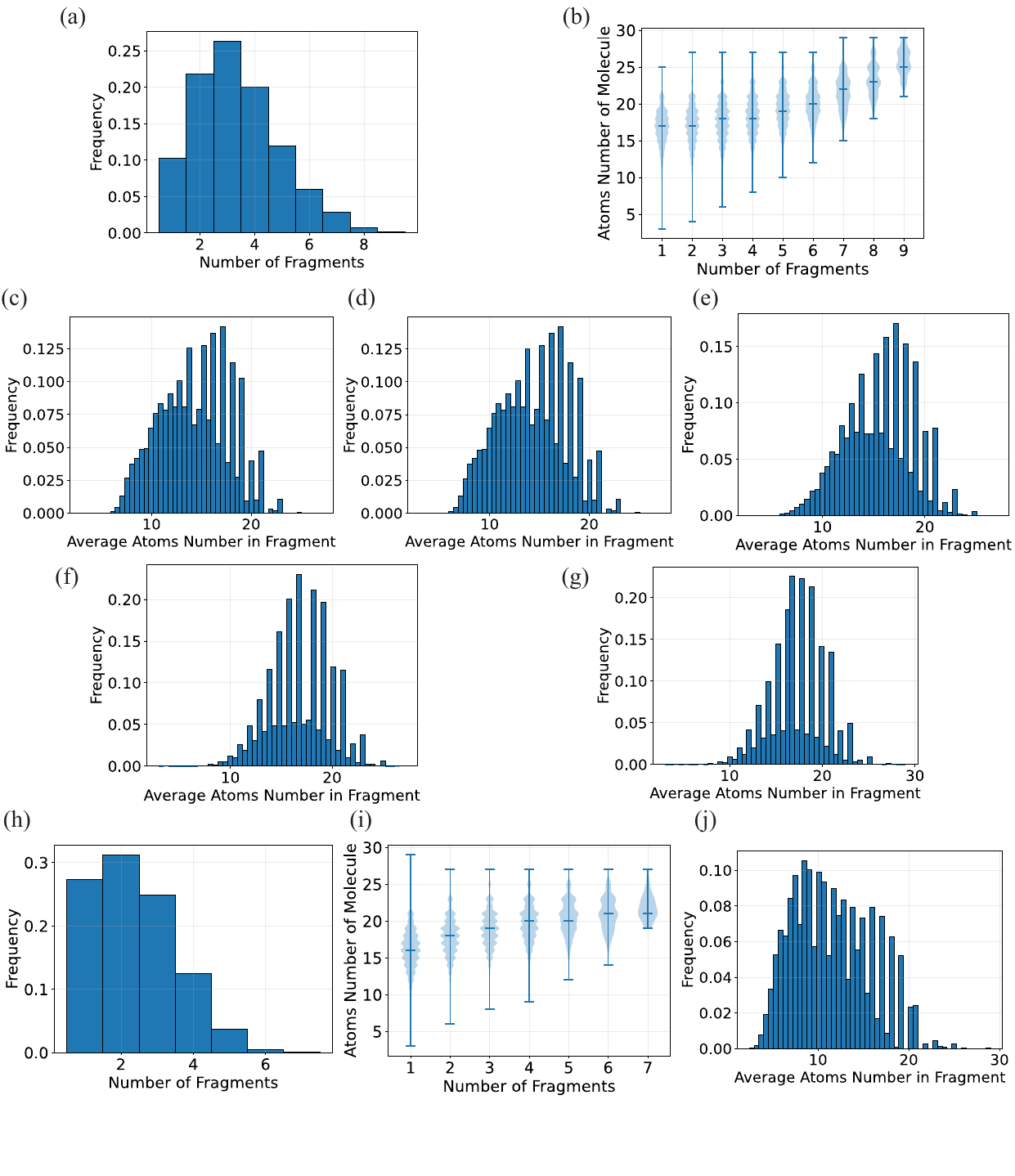}}
\caption{Distribution of fragments in QM9 dataset. (a) Fragment Count Distribution. The distribution remains consistent regardless of the value of $c_w$ or the bond order calculation method employed. (b) Molecule Size vs. Fragment Count Distribution. Generally, the more atoms molecule has, the more fragments will generate. It is also invariant for $c_w$ or bond order calculation scheme. Average Atom Count per Fragment Distribution (c) $c_w=0.1$, (d) $c_w=0.05$, (e) $c_w=0.01$ for Lendvay bond order and (f) $c_w=0.4$ and (g) $c_w=0.2$ for exponential bond order, respectively. (h) BRICS Fragment Count Distribution. (i) BRICS Molecule Size vs. Fragment Count Distribution (j) BRICS Average Atom Count per Fragment.}
\label{fig:frag_dis}
\end{center}
\end{figure}

\begin{table}[H]
\caption{Hyper-parameters for fragmentation. The expand threshold does not work for models training on MD22 dataset because they adapt \emph{implicit overlap} scheme.}
\label{tab:hyperparams-frag}
\begin{center}
\begin{small}
\begin{sc}
\resizebox{\linewidth}{!}{
\begin{tabular}{lcc}
\toprule
Bond Order Methods & Bond Order by Lendvay (\ref{eq:bo_lendvay}) & Frgmentation by Exponential (\ref{eq:bo_exp}) \\
\midrule
minimum atoms number $n_{\min}$ & 2 & 2 \\
maximum atoms number $n_{\max}$ & 6 & 6 \\
isolated threshold ($c_{is}$) & 0.1 & 0.4 \\
expand threshold ($c_w$) & 0.1 & 0.2, 0.4 \\
\bottomrule
\end{tabular}}
\end{sc}
\end{small}
\end{center}
\end{table}

\begin{table}[H]
\caption{Hyper-parameters for step \hyperref[enmt:expand_frag_implicit]{4*} of \emph{implicit overlap} scheme in MD22 experiments.}
\label{tab:hyperparams-frag-md22rc}
\begin{center}
\begin{small}
\begin{sc}
\resizebox{\linewidth}{!}{
\begin{tabular}{lccccc}
\toprule
 Molecules & Ac-Ala3-NHMe & DHA & Stachyose & AT-AT & AT-AT-CG-CG  \\
\midrule
distance cutoff $r_c$ (\AA) & 5.0 & 4.0 & 4.0 & 6.0 & 6.0\\
\bottomrule
\end{tabular}}
\end{sc}
\end{small}
\end{center}
\end{table}

\begin{table}[H]
\caption{Hyper-parameters for training SE3Set model. In the context of hyperparameter settings for dimensions, the symbols $e$ and $o$ are used to denote even and odd parity, respectively.}
\label{tab:hyperparams-model}
\begin{center}
\begin{small}
\begin{sc}
\resizebox{\linewidth}{!}{
\begin{tabular}{lccc}
\toprule
 & QM9 & MD17 & MD22\\
Hyper-parameters & Value or discriptions & Value or discriptions & Value or discriptions\\
\midrule
Optimizer $n_{\min}$ & AdamW & AdamW & AdamW\\
Learning rate scheduler & Cosine & Cosine & Cosine\\
Warm up epochs$n_{\max}$ & $5$ & $10$ & $10$\\
Minimum learning rate & $1.0\times 10^{-6}$ & $1.0\times 10^{-6}$ & $1.0\times 10^{-6}$\\
Batch size & $32,128$ & $8$ & $8$\\
Number of epochs & $400$ & $1500$ & $1500$\\
Weight decay & $5.0\times 10^{-3}$ & $1.0\times 10^{-6}$ & $1.0\times 10^{-6}$\\
Dropout rate & 0.1 & 0.0 & 0.0 \\
RBF cutoff (\AA) & 42.0 & \multicolumn{2}{c}{Max distance of used atom pairs}\\
Number of Basis & 128(Gaussian), 8(Bessel) & 32(Exponential)  & 32(Exponential)\\
Number of Blocks & 3, 6 &3, 6 & 6 \\
Node embedding dimension & \multicolumn{3}{c}{[(128, 0$e$), (64, 1$o$), (32, 2$e$)]}\\
Hyperedge embedding dimension & \multicolumn{3}{c}{[(128, 0$e$), (64, 1$o$), (32, 2$e$)]} \\
Attention head dimension & \multicolumn{3}{c}{[(32, 0$e$), (16, 1$o$), (8, 2$e$)]}\\
Feed forward dimension & \multicolumn{3}{c}{[(384, 0$e$), (192, 1$o$), (96, 2$e$)]}\\
Output feature dimension & \multicolumn{3}{c}{[(512, 0$e$)]}\\
\bottomrule
\end{tabular}}
\end{sc}
\end{small}
\end{center}
\end{table}


\newpage
\section*{NeurIPS Paper Checklist}

\begin{enumerate}

\item {\bf Claims}
    \item[] Question: Do the main claims made in the abstract and introduction accurately reflect the paper's contributions and scope?
    \item[] Answer: \answerYes{}
    \item[] Justification: The abstract and introduction have clearly stated the claims made, including the contributions made in the paper. The related material for the question can be found in abstract section.

\item {\bf Limitations}
    \item[] Question: Does the paper discuss the limitations of the work performed by the authors?
    \item[] Answer: \answerYes{}
    \item[] Justification: The related material for the question can be found in Sec.~\ref{sec:fragmentation} and Sec.~\ref{sec:md22}.

\item {\bf Theory Assumptions and Proofs}
    \item[] Question: For each theoretical result, does the paper provide the full set of assumptions and a complete (and correct) proof?
    \item[] Answer: \answerYes{}
    \item[] Justification: The related material for the question can be found in Sec.~\ref{sec:fragmentation} and Appendix.

    \item {\bf Experimental Result Reproducibility}
    \item[] Question: Does the paper fully disclose all the information needed to reproduce the main experimental results of the paper to the extent that it affects the main claims and/or conclusions of the paper (regardless of whether the code and data are provided or not)?
    \item[] Answer: \answerYes{}
    \item[] Justification: We have provided all the information needed to reproduce experimental results in Appendix~\ref{sec:training}. The code will be open source after the paper is accepted.

\item {\bf Open access to data and code}
    \item[] Question: Does the paper provide open access to the data and code, with sufficient instructions to faithfully reproduce the main experimental results, as described in supplemental material?
    \item[] Answer: \answerYes{}
    \item[] Justification: The code will be open source after the paper is accepted.

\item {\bf Experimental Setting/Details}
    \item[] Question: Does the paper specify all the training and test details (e.g., data splits, hyperparameters, how they were chosen, type of optimizer, etc.) necessary to understand the results?
    \item[] Answer: \answerYes{}
    \item[] Justification: We have provided all the information needed to reproduce experimental results in Appendix~\ref{sec:training}.

\item {\bf Experiment Statistical Significance}
    \item[] Question: Does the paper report error bars suitably and correctly defined or other appropriate information about the statistical significance of the experiments?
    \item[] Answer: \answerYes{}
    \item[] Justification: The mean and variance of the experimental results are reported in Fig.~\ref{fig:frag_dis}.

\item {\bf Experiments Compute Resources}
    \item[] Question: For each experiment, does the paper provide sufficient information on the computer resources (type of compute workers, memory, time of execution) needed to reproduce the experiments?
    \item[] Answer: \answerYes{}
    \item[] Justification: We have provided all the information in Appendix~\ref{sec:training}.
    
\item {\bf Code Of Ethics}
    \item[] Question: Does the research conducted in the paper conform, in every respect, with the NeurIPS Code of Ethics \url{https://neurips.cc/public/EthicsGuidelines}?
    \item[] Answer:\answerYes{}
    \item[] Justification: We have reviewed the NeurIPS Code of Ethics.

\item {\bf Broader Impacts}
    \item[] Question: Does the paper discuss both potential positive societal impacts and negative societal impacts of the work performed?
    \item[] Answer: \answerNA{}
    \item[] Justification: There is no societal impact of the work performed.
    
\item {\bf Safeguards}
    \item[] Question: Does the paper describe safeguards that have been put in place for responsible release of data or models that have a high risk for misuse (e.g., pretrained language models, image generators, or scraped datasets)?
    \item[] Answer: \answerNA{}
    \item[] Justification: The paper poses no such risks.

\item {\bf Licenses for existing assets}
    \item[] Question: Are the creators or original owners of assets (e.g., code, data, models), used in the paper, properly credited and are the license and terms of use explicitly mentioned and properly respected?
    \item[] Answer: \answerYes{}
    \item[] Justification: All the data can be found publicly, and the code will be open source after the paper is accepted. 

\item {\bf New Assets}
    \item[] Question: Are new assets introduced in the paper well documented and is the documentation provided alongside the assets?
    \item[] Answer: \answerNA{}
    \item[] Justification: The paper does not release new assets.

\item {\bf Crowdsourcing and Research with Human Subjects}
    \item[] Question: For crowdsourcing experiments and research with human subjects, does the paper include the full text of instructions given to participants and screenshots, if applicable, as well as details about compensation (if any)? 
    \item[] Answer: \answerNA{}
    \item[] Justification: The paper does not involve crowdsourcing nor research with human subjects.

\item {\bf Institutional Review Board (IRB) Approvals or Equivalent for Research with Human Subjects}
    \item[] Question: Does the paper describe potential risks incurred by study participants, whether such risks were disclosed to the subjects, and whether Institutional Review Board (IRB) approvals (or an equivalent approval/review based on the requirements of your country or institution) were obtained?
    \item[] Answer: \answerNA{}
    \item[] Justification: The paper does not involve crowdsourcing nor research with human subjects.

\end{enumerate}

\end{document}